\documentclass[twoside,11pt]{article}

\usepackage[preprint]{jmlr2e}
\usepackage[english]{babel}
\usepackage{color}
\usepackage{xcolor}
\usepackage{url}
\usepackage{booktabs}
\usepackage{amsfonts,amssymb}
\usepackage{amsmath}
\usepackage{amsthm}
\usepackage{bbding}
\usepackage{pifont}
\newcommand{\cmark}{\ding{51}}
\newcommand{\xmark}{\ding{55}}
\usepackage{wasysym}
\usepackage{utfsym}
\usepackage{fontawesome}
\usepackage{colortbl}
\usepackage{bm}
\usepackage{hyperref}
\usepackage{tikz}
\usetikzlibrary{positioning}
\usetikzlibrary{mindmap,trees}
\definecolor{mygray}{gray}{.9}
\usepackage{tikz}
\usetikzlibrary{trees}
\usepackage{microtype}      
\usepackage{graphicx} 
\usepackage{listings}
\usepackage{subcaption}
\usepackage{longtable}
\usepackage{multirow}
\hypersetup{
    colorlinks=true,
    linkcolor=red,
    citecolor=blue,
    filecolor=magenta,      
    urlcolor=blue,
linktocpage}

\definecolor{codegreen}{rgb}{0,0.6,0}
\definecolor{codegray}{rgb}{0.5,0.5,0.5}
\definecolor{codepurple}{rgb}{0.58,0,0.82}
\definecolor{backcolour}{gray}{0.9}
\definecolor{color1}{HTML}{7FB7D8}
\definecolor{color2}{HTML}{D3B7D6}

\lstdefinestyle{mystyle}{
    backgroundcolor=\color{backcolour},   
    commentstyle=\color{codegreen},
    keywordstyle=\color{magenta},
    numberstyle=\tiny\color{codegray},
    stringstyle=\color{codepurple},
    basicstyle=\ttfamily\footnotesize,
    breakatwhitespace=false,         
    breaklines=true,                 
    captionpos=b,                    
    keepspaces=true,                 
    numbers=left,                    
    numbersep=5pt,                  
    showspaces=false,                
    showstringspaces=false,
    showtabs=false,                  
    tabsize=2
}
\newcommand{\omni}{\texttt{OmniSafe} }
\newcommand{\sg}{\texttt{Safety-Gymnasium} }
\newcommand{\ts}{\texttt{Tianshou} }
\newcommand{\stable}{\texttt{Stable-Baselines3} }

\newcommand{\clean}{\texttt{CleanRL} }
\newcommand{\safety}{\texttt{Safety-starter-agents} }
\newcommand{\ray}{\texttt{Ray/RLlib} }
\usepackage{tikz}
\usetikzlibrary{shapes}

\usepackage{lastpage}
\jmlrheading{23}{2022}{1-\pageref{LastPage}}{1/21; Revised 5/22}{9/22}{21-0000}{Jiaming Ji, Jiayi Zhou, Borong Zhang, Juntao Dai, Xuehai Pan, Ruiyang Sun, Weidong Huang, Yiran Geng, Mickel Liu, Yaodong Yang}

\ShortHeadings{OmniSafe: An Infrastructure for Accelerating Safe Reinforcement Learning Research}{Ji, Zhou, Zhang, Dai, Pan, Sun, Huang, Geng, Liu, Yang}
\firstpageno{1}
\begin{document}

\title{OmniSafe: An Infrastructure for Accelerating 

Safe Reinforcement Learning Research}

\author{\name Jiaming Ji$^{1}$\thanks{Jiaming Ji, Jiayi Zhou, and Borong Zhang are core developers. They contributed equally to this paper.}  \email jiamg.ji@gmail.com\\
       \name Jiayi Zhou$^{1}${\color{red}\footnotemark[1]} \email gaiejj@outlook.com \\
       \name Borong Zhang$^{1}${\color{red}\footnotemark[1]} \email borongzh@gmail.com \\
       \name Juntao Dai$^{1}$ \email jtd.acad@gmail.com\\
       \name Xuehai Pan$^{1}$ \email xuehaipan@pku.edu.cn\\
       \name Ruiyang Sun$^{1}$ \email sun\_ruiyang@stu.pku.edu.cn\\
       \name Weidong Huang$^{1}$ \email bigeasthuang@gmail.com\\
       \name Yiran Geng$^{1}$ \email gyr@stu.pku.edu.cn\\
       \name Mickel Liu$^{1}$ \email mickelliu7@gmail.com\\
       \name Yaodong Yang$^{1}$\thanks{Yaodong Yang is the corresponding author.\vspace{-0.55cm}} \email yaodong.yang@pku.edu.cn\\
       \addr $^{1}$ Peking University, China\\
}
\editor{My editor}

\maketitle

\begin{abstract}
AI systems empowered by reinforcement learning (RL) algorithms harbor the immense potential to catalyze societal advancement, yet their deployment is often impeded by significant safety concerns. Particularly in safety-critical applications, researchers have raised concerns about unintended harms or unsafe behaviors of unaligned RL agents. 
The philosophy of safe reinforcement learning (SafeRL) is to align RL agents with harmless intentions and safe behavioral patterns.
In SafeRL, agents learn to develop optimal policies by receiving feedback from the environment, while also fulfilling the requirement of minimizing the risk of unintended harm or unsafe behavior.
However, due to the intricate nature of SafeRL algorithm implementation, combining methodologies across various domains presents a formidable challenge. This had led to an absence of a cohesive and efficacious learning framework within the contemporary SafeRL research milieu.
In this work, we introduce a foundational framework designed to expedite SafeRL research endeavors. Our comprehensive framework encompasses an array of algorithms spanning different RL domains and places heavy emphasis on safety elements. Our efforts are to make the SafeRL-related research process more streamlined and efficient, therefore facilitating further research in AI safety. Our project is released at: \url{https://github.com/PKU-Alignment/omnisafe}.
\end{abstract}


\section{Introduction}
Reinforcement learning has gained immense attention as a powerful class of machine learning algorithms capable of addressing complex problems in diverse domains such as robotics and game playing \citep{go2016,go2017,alphastar}. Nevertheless, the application of reinforcement learning in safety-critical domains raises concerns about unintended consequences and potential harms \citep{human_ai_safe1,human_ai_safe2}. The self-governed learning of reinforcement learning agents, relying on environmental feedback, can give rise to unforeseen, unsafe control policies. This may result in adverse outcomes, as seen in autonomous vehicles valuing speed over safety. These concerns have prompted extensive research into developing safe reinforcement learning algorithms for safety-critical applications \citep{av_safe_nature}. 

\paragraph{Lack of OSS Infrastructure for SafeRL Research}
Despite the existence of open-source software (OSS) and frameworks, a unified and comprehensive code framework remains elusive in the SafeRL field. OpenAI introduced the safety-starter-agent \citep{safety_gym}, a SafeRL framework in 2019 that implemented four classical algorithms based on TensorFlow-v1. While this framework has aided subsequent researchers, it has not been updated or maintained since its release, and its core dependency TensorFlow-v1 has been deprecated\footnote{\url{https://github.com/openai/safety-starter-agents}} for years. 
SafeRL algorithm implementation is complex in nature such that the cross-domain integration with RL algorithms (\textit{i.e.} constraint optimization, safe control theory) poses a significant technical challenge, resulting in the ongoing absence of a unified and effective learning framework within the contemporary SafeRL research landscape.
This lack of a unified backend and extensible framework substantially impedes progress in the domain. 

To fulfill this gap, we present \omni, an infrastructural framework designed to accelerate SafeRL research by providing a wide range of algorithms, including Off-Policy, On-Policy, Model-based, and Offline algorithms, \textit{etc.} 

\section{Features of \omni}
The \omni framework is featured with the following contributions:
\paragraph{(1) High Modularity.} \omni is a highly modular framework, incorporating an extensive collection of dozens of algorithms tailored for SafeRL across diverse domains. The framework achieves versatility through the utilization of algorithm-level abstraction and API, employing two distinctive design components: \texttt{Adapter} and \texttt{Wrapper}. The \texttt{Adapter} component serves to reconcile structural differences between algorithms for various classes of Constrained Markov Decision Processes (CMDPs)~\citep{cmdp, ET-mdp, sauteRL2022}, while the \texttt{Wrapper} component is designed to mitigate the variability (\textit{e.g.} dynamics) inherent in algorithm-environment interactions. 

\paragraph{(2) High-performance parallel computing acceleration.} By harnessing the capabilities of \texttt{torch.distributed}, \omni accelerates the learning process of algorithms with process parallelism. This enables \omni not only to support environment-level asynchronous parallelism but also incorporates agent asynchronous learning. This methodology bolsters training stability and expedites the training process via the deployment of a parallel exploration mechanism. The integration of agent asynchronous learning in \omni underscores its commitment to providing a versatile and robust platform for advancing SafeRL research.

\paragraph{(3) Code reliability and reproducibility.} The algorithms implemented \omni have been rigorously tested in Safety-Gym~\citep{safety_gym} and Mujoco-Velocity~\citep{focops2020} environments to confirm their consistency with the results presented in the original papers. Code examples necessary for result reproduction are provided within the project. Furthermore, observations, analyses, and raw data from this process will be made accessible to the community as a valuable reference. To maintain stringent code quality, \omni's code has undergone thorough unit testing throughout the development process and adheres to project management standards such as \texttt{GitHub Actions CI}, \texttt{Pylint}, and \texttt{MyPy}.

\paragraph{(4) Fostering the Growth of SafeRL Communitiy.} \omni is a unified learning framework that represents a significant advance in standardizing the field of SafeRL. 
The platform offers comprehensive API documentation, including user guides describing both fundamental and advanced features, illustrative examples, and best practices for algorithm parameter selection and result monitoring. Moreover, the documentation delves into detailed theoretical algorithm derivations alongside practical tutorials, thereby facilitating a swift grasp of the SafeRL algorithm for novices.

\begin{figure}[hbtp]
  \centering
  \includegraphics[width=1.0\linewidth]{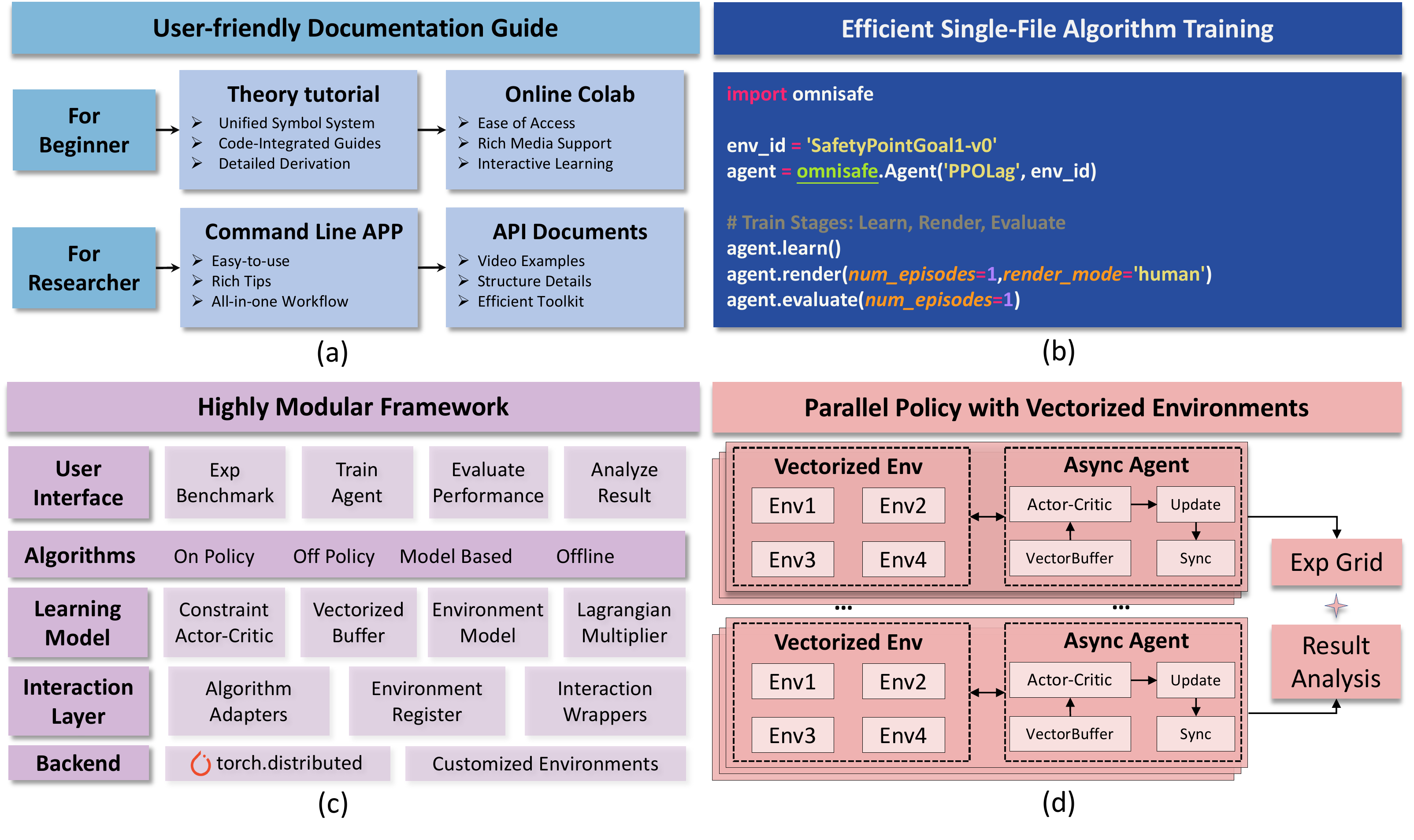}
  \caption{The core features of \omni include (a) Comprehensive API documentation with user guides, examples, and best practices for efficient learning, the documentation can be found in \url{https://omnisafe.readthedocs.io}; (b) Streamlined algorithm training through single-file execution, simplifying setup and management. (c) Achieve versatility through the utilization of algorithm-level abstraction and API interfaces; (d) Enhanced training stability and speed with environment-level asynchronous parallelism and agent asynchronous learning.}
  \vspace{-2em}
  \label{framework}
\end{figure}

\section{DataFlows of \omni}
\begin{figure}[hbtp]
  \centering
  \includegraphics[width=1.0\linewidth]{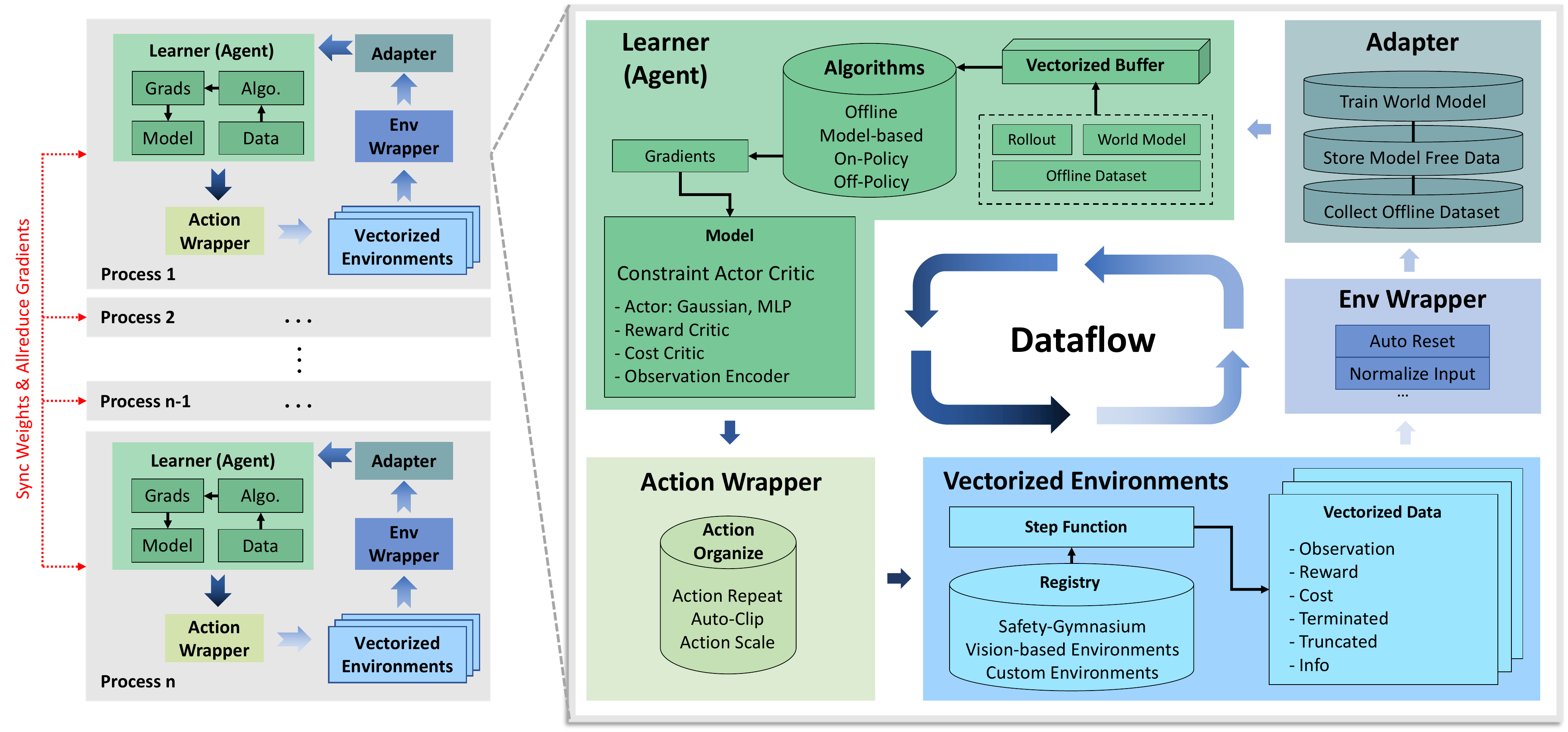}
  \caption{A high-level depiction of \omni’s distributed dataflow process. Each process periodically syncs weights and all-reduce gradients with other processes. \texttt{Vectorized Environments} first generate trajectories of the agent's interactions with the environment. Second, the \texttt{EnvWrapper} monitors and governs the environment's status (\textit{e.g.} Auto-Reset) and outputs. Then, the \texttt{Adapter} assigns a suitable execution plan that handles data pre-processing. Next, the \texttt{Learner} gathers pre-processed data, calls the learning algorithm, and trains the model. Lastly, the \texttt{ActionWrapper} transforms the model's outputs to the agent's actions interpretable by the environments. Thereby completing a cycle of dataflow.
}
\vspace{-2em}
  \label{dataflow}
\end{figure}

To accommodate the rapid development of various algorithms and mitigate incompatibilities across different environment interfaces, 
we have devised a unified dataflow interaction framework for \omni as illustrated in \ref{dataflow}.  
The utilization of \texttt{Adapter} and \texttt{Wrapper} designs in \omni facilitates enhanced reusability and compatibility with current code resources, thereby reducing the need for additional engineering efforts when creating new algorithms or integrating new environments.
A key highlight is that the \texttt{Adapter} layer also provides a convenient approach for transforming problem paradigms and extracting information from the data inflow.
For instance, to convert the CMDP \citep{cmdp} paradigm to the SauteMDP \citep{sauteRL2022} paradigm, one simply needs to replace the \texttt{Adapter} on the inflow side of data with that of the SauteMDP paradigm. This approach enables users to concentrate more on the problem at hand while incurring minimal code alterations, by utilizing the modular tools provided by \omni.

\section{Conclusion and Outlook}
In conclusion, \omni provides a comprehensive infrastructural framework designed to expedite SafeRL research, featuring extensive API documentation, compatibility with popular environments, high modularity, and support for distributive computing. 
We have experimentally validated the implementation of numerous algorithms spanning diverse RL domains, facilitating efficient experimentation and verification of novel concepts. 
Our documentation delivers practical advice for conducting SafeRL research and supplies a developer guide for enhancing the platform's functionality.

Our endeavors are directed toward the standardization of SafeRL-related research tools and methods, with the aim of cultivating a more streamlined and highly efficient approach. By doing so, we endeavor to facilitate and expedite further scientific exploration in the field of AI safety.

\clearpage
\bibliography{sample}

\newpage

\appendix

\section*{Appendix A.}

\section{Implemented Algorithms of \omni}

\omni offers a highly modular framework that integrates an extensive collection of algorithms specifically designed for Safe Reinforcement Learning (SafeRL) in various domains. The \texttt{Adapter} module in \omni allows for easy expansion of different types of SafeRL algorithms.

\begin{table}[htb]
\resizebox{\columnwidth}{!}{
\begin{tabular}{@{}l|l|l@{}}
\toprule
\textbf{Domains}                  & \textbf{Types}       & \textbf{Algorithms Registry}                                                     \\ \midrule
\multirow{4}{*}{\textbf{On-Policy}} &
  Primal-Dual &
  \begin{tabular}[c]{@{}l@{}}TRPO-Lag\citep{safety_gym}; PPO-Lag\citep{safety_gym}; TRPO-PID\citep{PIDLag2020};\\ CPPO-PID\citep{PIDLag2020}; PDO\citep{pdo}; RCPO\citep{rcpo2018}; \end{tabular} \\ \cmidrule(l){2-3} 
                                  & Primal               & OnCRPO\citep{crpo};                                                                   \\ \cmidrule(l){2-3} 
                                  & Convex Optimization  & 
                                   \begin{tabular}[c]{@{}l@{} }CPO\citep{cpo2017}; PCPO\citep{pcpo2020}; 
                                   CUP\citep{cup2022}; \\    FOCOPS\citep{focops2020} ;\end{tabular}                                             \\ \cmidrule(l){2-3} 
                                  & Penalty Function     & IPO\citep{ipo2020}; P3O\citep{p3o2022};                                                      \\ \midrule
\textbf{Off Policy}               & Primal-Dual          & SAC-Lag\citep{safety_gym}; DDPG-Lag\citep{safety_gym};  TD3-Lag\citep{safety_gym};                             \\ \midrule
\multirow{3}{*}{\textbf{Model-based}} &
  Online Plan &
  SafeLOOP\citep{loop2022}; CCEPETS\citep{ccme2018}; RCEPETS\citep{rce}; \\ \cmidrule(l){2-3} 
                                  & Pessimistic Estimate & LA-MBDA\citep{lambda2022}; CAPPETS\citep{cap2022};                                              \\ \cmidrule(l){2-3} 
                                  & Imaginary Train      & SMBPO\citep{smbpo2021}; MBPPO-Lag\citep{mbppolag2022};                                               \\ \midrule
\multirow{2}{*}{\textbf{Offline}} & Q-Learning Based     & BCQLag\citep{bcq2019}; C-CRR\citep{crr2020}; VAE-BC\citep{bcq2019};                                               \\ \cmidrule(l){2-3} 
                                  & DICE Based           & COptiDICE\citep{coptidice2022};                                                               \\ \midrule
\multirow{3}{*}{\textbf{\begin{tabular}[c]{@{}l@{}}Other Formulation\\ MDP\end{tabular}}} &
  ET-MDP &
  PPOEarlyTerminated\citep{ET-mdp};  TRPOEarlyTerminated\citep{ET-mdp}; \\ \cmidrule(l){2-3} 
                                  & SauteRL              & \begin{tabular}[c]{@{}l@{}}PPOSaute\citep{sauteRL2022}; TRPOSaute\citep{sauteRL2022};\end{tabular} \\ \cmidrule(l){2-3} 
                                  & SimmerRL             & PPOSimmerPID\citep{swimmer2022}; TRPOSimmerPID\citep{swimmer2022};                                   \\ \bottomrule
\end{tabular}
}
\caption{\omni supports varieties of SafeRL algorithms. From the perspective of classic reinforcement learning, \omni includes on-policy, off-policy, offline, and model-based algorithms; From the perspective of the SafeRL learning paradigm, \omni supports primal-dual, projection, penalty function, primal, etc.}
\end{table}

\section{\sg Experiment Results of \omni}
\begin{figure}[htb]
  \centering
  \begin{subfigure}{0.3\linewidth}
    \centering
    \includegraphics[width=\linewidth]{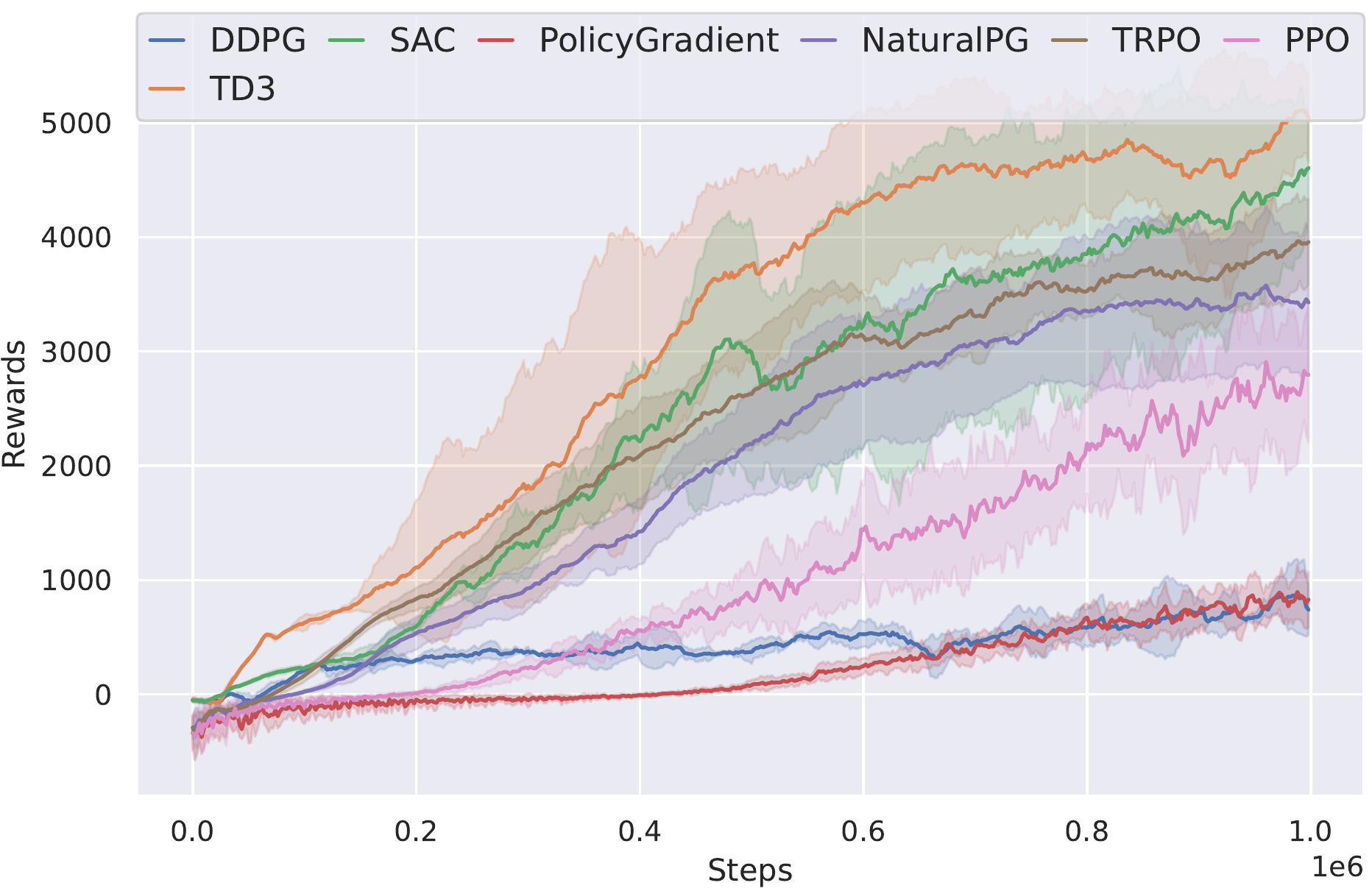}
    \caption{Ant}
  \end{subfigure}
  \begin{subfigure}{0.3\linewidth}
    \centering
    \includegraphics[width=\linewidth]{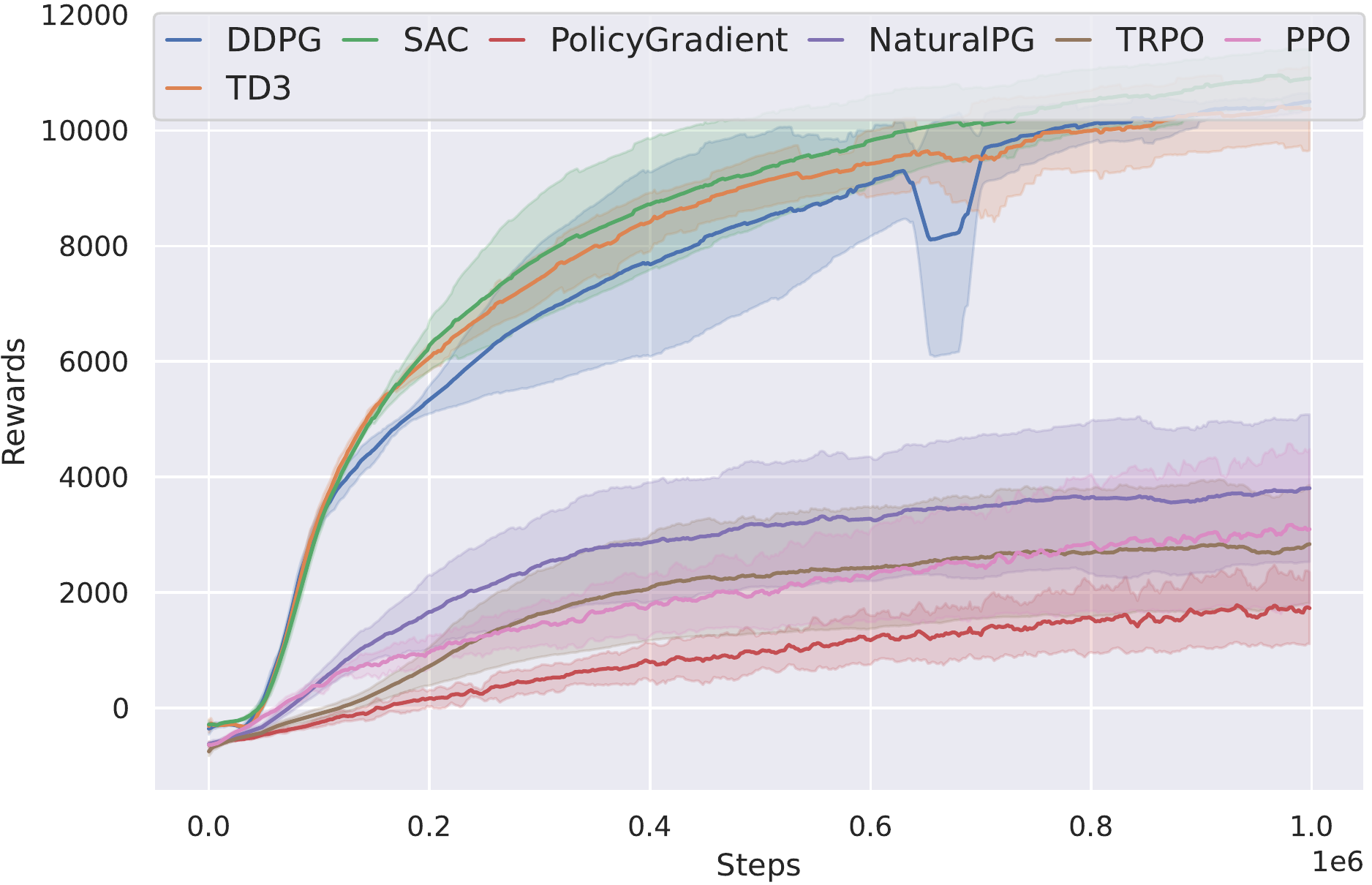}
    \caption{HalfCheetah}
  \end{subfigure}
  \begin{subfigure}{0.3\linewidth}
    \centering
    \includegraphics[width=\linewidth]{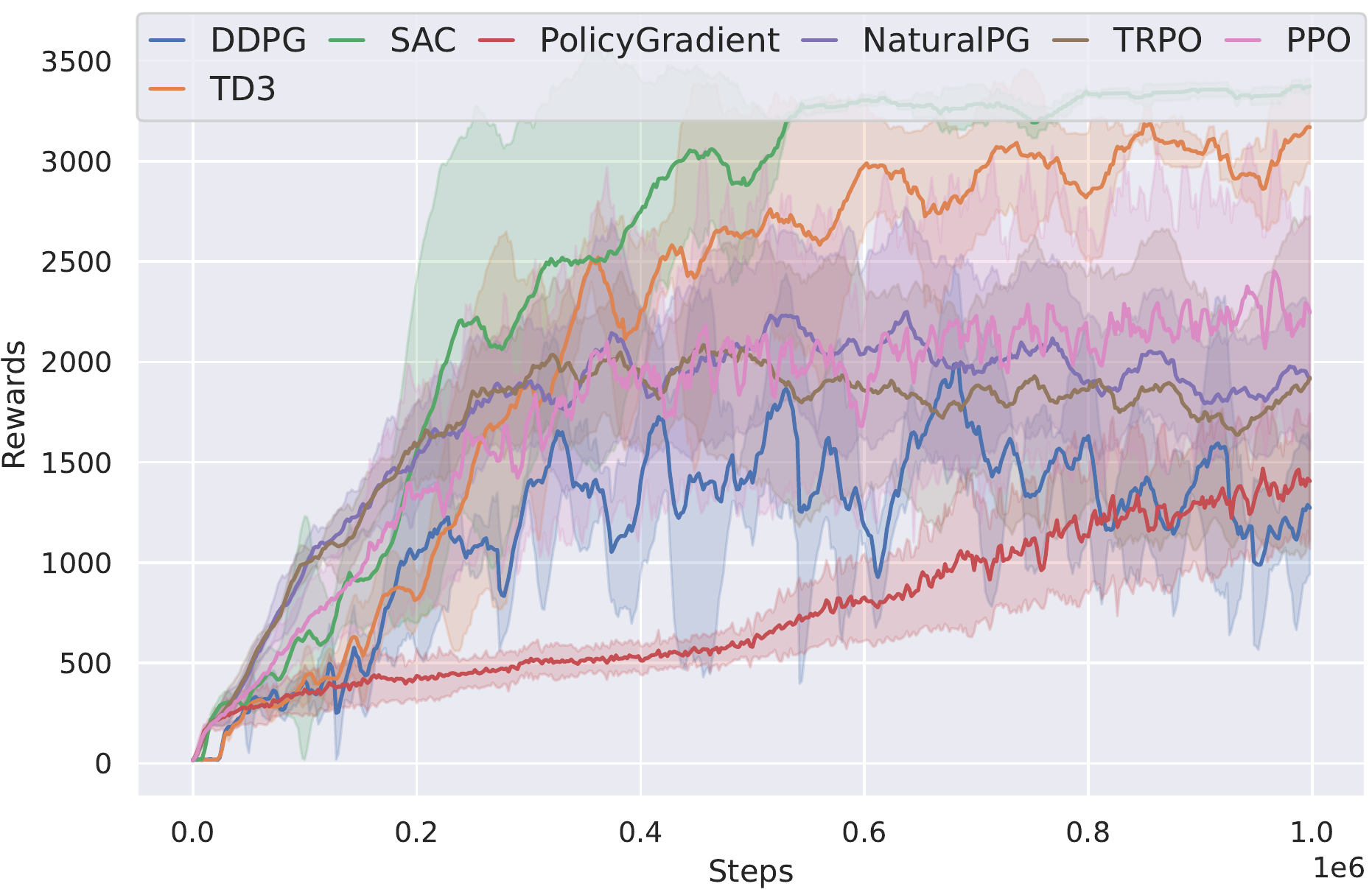}
    \caption{Hopper}
  \end{subfigure}
  \begin{subfigure}{0.3\linewidth}
    \centering
    \includegraphics[width=\linewidth]{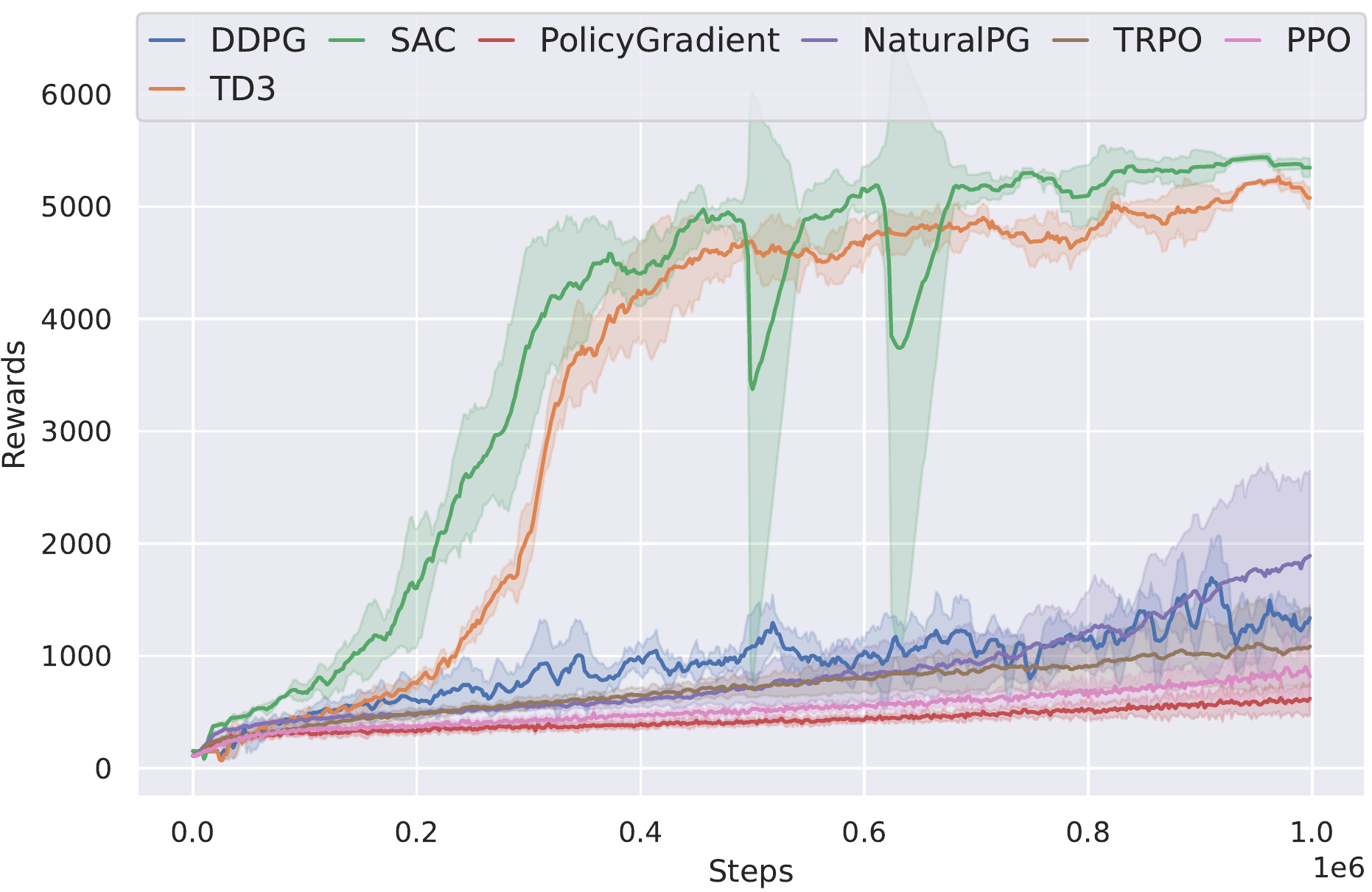}
    \caption{Humanoid}
  \end{subfigure}
  \begin{subfigure}{0.3\linewidth}
    \centering
    \includegraphics[width=\linewidth]{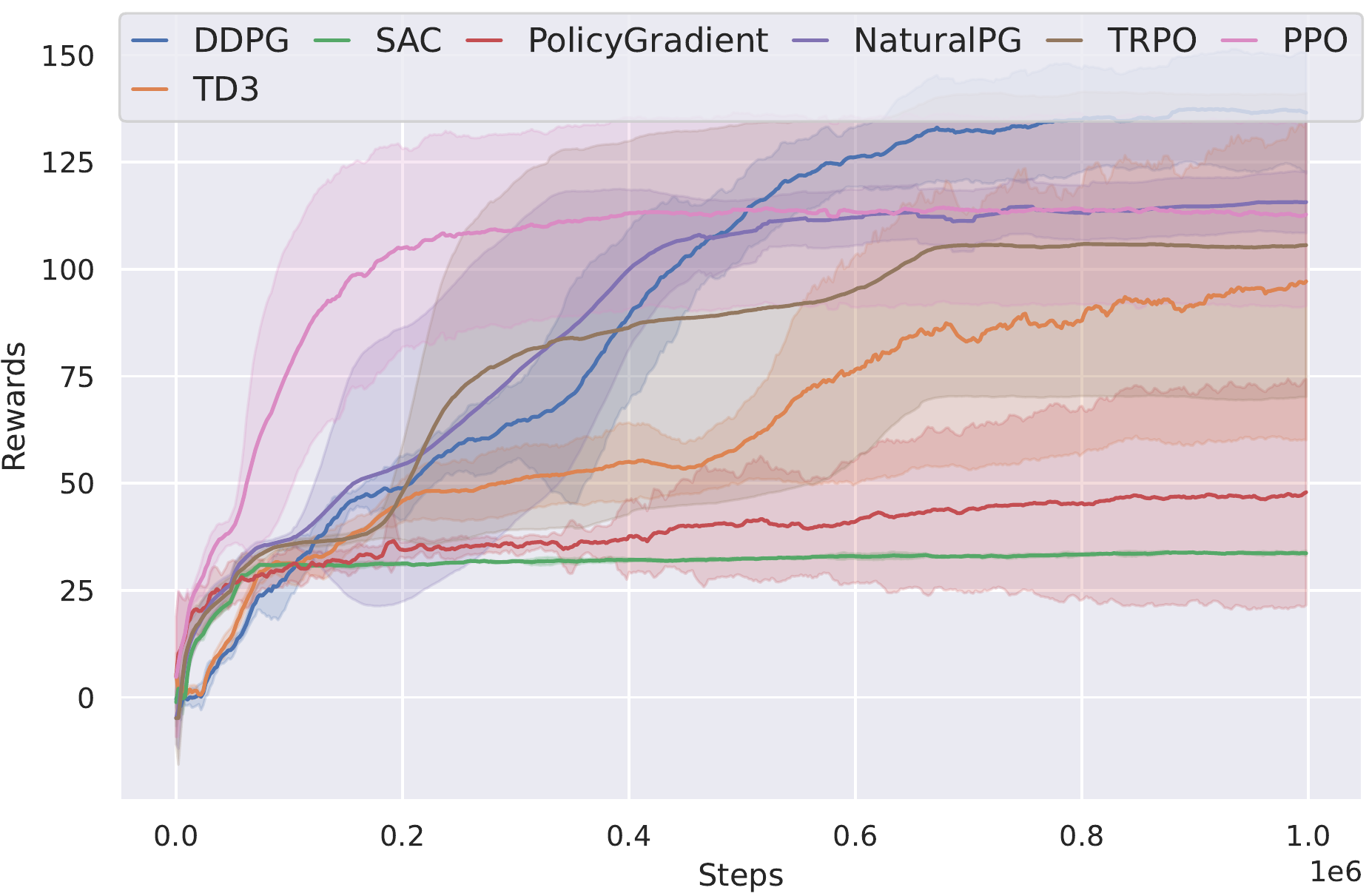}
    \caption{Swimmer}
  \end{subfigure}
  \begin{subfigure}{0.3\linewidth}
    \centering
    \includegraphics[width=\linewidth]{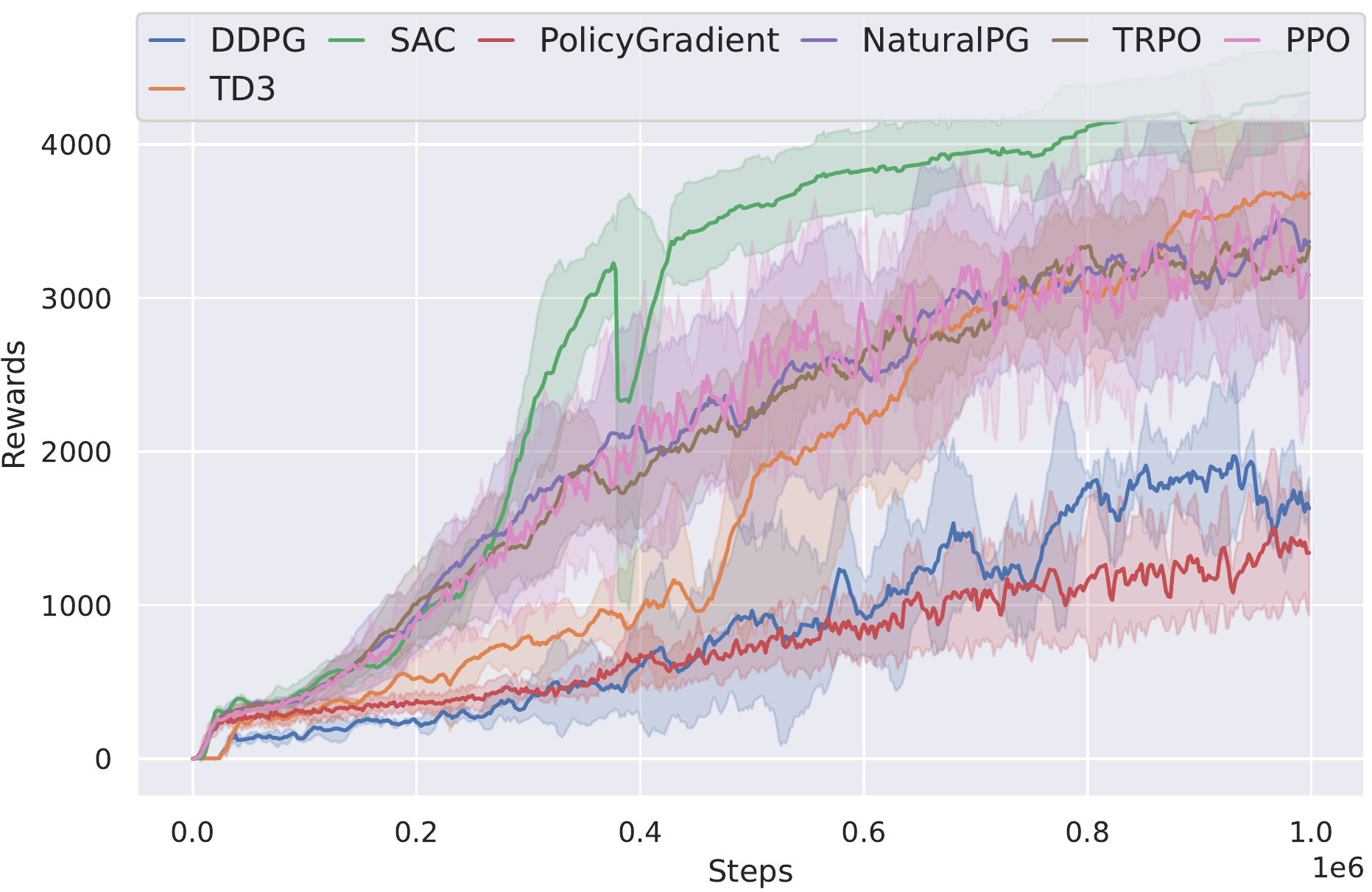}
    \caption{Walker2d}
  \end{subfigure}
\caption{Training curves in \sg MuJoCo Velocity environments, covering all classical reinforcement learning algorithms mentioned in \autoref{compare}. The rewards are obtained from the \texttt{1e6} steps interaction.}
\label{1e6}
\end{figure}

In an effort to ascertain the credibility of \omni's algorithmic implementation, a comparative assessment was conducted, juxtaposing the performance of classical reinforcement learning algorithms, such as Policy Gradient \citep{pg1999}, Natural Policy Gradient \citep{npg2001}, TRPO \citep{trpo2015}, PPO \citep{ppo2017}, DDPG \citep{ddpg2015}, TD3 \citep{td32018}, and SAC \citep{sac2018}, with well-established open-source implementations, specifically \ts \citep{tianshou2022} \footnote{The Project URL: \url{https://github.com/thu-ml/tianshou/tree/master}} and \stable \citep{sb32021}\footnote{The Project URL: \url{https://github.com/DLR-RM/stable-baselines3}}. The stringent verification procedure undertaken has yielded confidence in the reliability of \omni's algorithms. The comparison results with \ts and \stable are shown in \autoref{compare} . The specific performance curve of \omni on \sg Mujoco Velocity is shown in \autoref{1e6}

\begin{table}[htb]

\captionsetup[subtable]{justification=centering}

\begin{subtable}{\linewidth}\centering
\resizebox{\columnwidth}{!}{
\begin{tabular}{@{}l|ccc|ccc@{}}\toprule
    & \multicolumn{3}{c|}{\textbf{Policy Gradient}}  & \multicolumn{3}{c}{\textbf{PPO}} \\
\midrule
\textbf{Environment} \hfill  & \textbf{OmniSafe (Ours)} & \textbf{Tianshou} & \textbf{Stable-Baselines3} & \textbf{OmniSafe (Ours)} & \textbf{Tianshou} & \textbf{Stable-Baselines3}   \\ \midrule
\textsc{SafetyAntVelocity-v1} & \textbf{2769.45 $\pm$ 550.71} & 145.33 $\pm$ 127.55 & - $\pm$ - & \textbf{4295.96 $\pm$ 658.20} & 2607.48 $\pm$ 1415.78 & 1780.61 $\pm$ 780.65  \\
\textsc{SafetyHalfCheetahVelocity-v1} & \textbf{2625.44 $\pm$ 1079.04} & 707.56 $\pm$ 158.59 & - $\pm$ - & 3507.47 $\pm$ 1563.69 & \textbf{6299.27 $\pm$ 1692.38} & 5074.85 $\pm$ 2225.47  \\
\textsc{SafetyHopperVelocity-v1} & \textbf{1884.38 $\pm$ 825.13} & 343.88 $\pm$ 51.85 & - $\pm$ - & \textbf{2679.98 $\pm$ 921.96} & 1834.70 $\pm$ 862.06 & 838.96 $\pm$ 351.10  \\
\textsc{SafetyHumanoidVelocity-v1} & \textbf{647.52 $\pm$ 154.82} & 438.97 $\pm$ 123.68 & - $\pm$ - & \textbf{1106.09 $\pm$ 607.6} & 677.43 $\pm$ 189.96 & 762.73 $\pm$ 170.22  \\
\textsc{SafetySwimmerVelocity-v1} & \textbf{47.31 $\pm$ 16.19} & 27.12 $\pm$ 7.47 & - $\pm$ - & 113.28 $\pm$ 20.22 & 37.93 $\pm$ 8.68 & \textbf{273.86 $\pm$ 87.76}  \\
\textsc{SafetyWalker2dVelocity-v1} & \textbf{1665 .00 $\pm$ 930.18} & 373.63 $\pm$ 129.20 & - $\pm$ - & \textbf{3806.39 $\pm$ 1547.48} & 3748.26 $\pm$ 1832.83 & 3304.35 $\pm$ 706.13  \\
\bottomrule
\end{tabular}
}
\end{subtable}

\begin{subtable}{\linewidth}\centering
\resizebox{\columnwidth}{!}{
\begin{tabular}{@{}l|ccc|ccc@{}}\toprule
    & \multicolumn{3}{c|}{\textbf{Natural PG}}  & \multicolumn{3}{c}{\textbf{TRPO}} \\
\midrule
\textbf{Environment} \hfill  & \textbf{OmniSafe (Ours)} & \textbf{Tianshou} & \textbf{Stable-Baselines3} & \textbf{OmniSafe (Ours)} & \textbf{Tianshou} & \textbf{Stable-Baselines3}   \\ \midrule
\textsc{SafetyAntVelocity-v1} & \textbf{3793.70 $\pm$ 583.66} & 2062.45 $\pm$ 876.43 & - $\pm$ - & \textbf{4362.43 $\pm$ 640.54} & 2521.36 $\pm$ 1442.10 & 3233.58 $\pm$ 1437.16  \\
\textsc{SafetyHalfCheetahVelocity-v1} & \textbf{4096.77 $\pm$ 1223.70} & 3430.90 $\pm$ 239.38 & - $\pm$ - & 3313.31 $\pm$ 1048.78 & 4255.73 $\pm$ 1053.82 & \textbf{7185.06 $\pm$ 3650.82}  \\
\textsc{SafetyHopperVelocity-v1} & \textbf{2590.54 $\pm$ 631.05} & 993.63 $\pm$ 489.42 & - $\pm$ - & \textbf{2698.19 $\pm$ 568.80} & 1346.94 $\pm$ 984.09 & 2467.10 $\pm$ 1160.25  \\
\textsc{SafetyHumanoidVelocity-v1} & \textbf{3838.67 $\pm$ 1654.79} & 810.76 $\pm$ 270.69 & - $\pm$ - & 1461.51 $\pm$ 602.23 & 749.42 $\pm$ 149.81 & \textbf{2828.18 $\pm$ 2256.38}  \\
\textsc{SafetySwimmerVelocity-v1} & \textbf{116.33 $\pm$ 5.97} & 29.75 $\pm$ 12.00 & - $\pm$ - & 105.08 $\pm$ 31.00 & 37.21 $\pm$ 4.04 & \textbf{258.62 $\pm$ 124.91}  \\
\textsc{SafetyWalker2dVelocity-v1} & \textbf{4054.62 $\pm$ 1266.76} & 3372.59 $\pm$ 1049.14 & - $\pm$ - & 4099.97 $\pm$ 409.05 & 3372.59 $\pm$ 961.74 & \textbf{4227.91 $\pm$ 760.93}  \\
\bottomrule
\end{tabular}
}
\end{subtable}

\begin{subtable}{\linewidth}\centering
\resizebox{\columnwidth}{!}{
\begin{tabular}{@{}l|ccc|ccc|ccc@{}}\toprule
    & \multicolumn{3}{c|}{\textbf{DDPG}}  & \multicolumn{3}{c|}{\textbf{TD3}} & \multicolumn{3}{c}{\textbf{SAC}}\\
\midrule
\textbf{Environment} \hfill  & \textbf{OmniSafe (Ours)} & \textbf{Tianshou} & \textbf{Stable-Baselines3} & \textbf{OmniSafe (Ours)} & \textbf{Tianshou} & \textbf{Stable-Baselines3} & \textbf{OmniSafe (Ours)} & \textbf{Tianshou} & \textbf{Stable-Baselines3}  \\ \midrule
\textsc{SafetyAntVelocity-v1} & 860.86 $\pm$ 198.03 & 308.60 $\pm$ 318.60 &\textbf{ 2654.58 $\pm$ 1738.21} & 5246.86 $\pm$ 580.50 & \textbf{5379.55 $\pm$ 224.69} & 3079.45 $\pm$ 1456.81  & 5456.31 $\pm$ 156.04 & \textbf{6012.30 $\pm$ 102.64} & 2404.50 $\pm$ 1152.65\\
\textsc{SafetyHalfCheetahVelocity-v1} & 11377.10 $\pm$ 75.29 & \textbf{12493.55 $\pm$ 437.54} & 7796.63 $\pm$ 3541.64 & \textbf{11246.12 $\pm$ 488.62} & 10246.77 $\pm$ 908.39 & 8631.27 $\pm$ 2869.15  & 11488.86 $\pm$ 513.09 & \textbf{12083.89 $\pm$ 564.51} & 7767.74 $\pm$ 3159.07\\
\textsc{SafetyHopperVelocity-v1} & 1462.56 $\pm$ 591.14 & 2018.97 $\pm$ 1045.20 & \textbf{2214.06 $\pm$ 1219.57} & \textbf{3404.41 $\pm$ 82.57} & 2682.53 $\pm$ 1004.84 & 2542.67 $\pm$ 1253.33  & \textbf{3597.70 $\pm$ 32.23} & 3546.59 $\pm$ 76 .00 & 2158.54 $\pm$ 1343.24\\
\textsc{SafetyHumanoidVelocity-v1} & 1537.39 $\pm$ 335.62 & 124.96 $\pm$ 61.68 & \textbf{2276.92 $\pm$ 2299.68} & \textbf{5798.01 $\pm$ 160.7}2 & 3838.06 $\pm$ 1832.90 & 3511.06 $\pm$ 2214.12  & \textbf{6039.77 $\pm$ 167.82} & 5424.55 $\pm$ 118.52 & 2713.60 $\pm$ 2256.89\\
\textsc{SafetySwimmerVelocity-v1} & 139.39 $\pm$ 11.74 & 138.98 $\pm$ 8.60 & \textbf{210.40 $\pm$ 148.01} & \textbf{98.39 $\pm$ 32.28} & 94.43 $\pm$ 9.63 & 247.09 $\pm$ 131.69  & 46.44 $\pm$ 1.23 & 44.34 $\pm$ 2.01 & \textbf{247.33 $\pm$ 122.02}\\
\textsc{SafetyWalker2dVelocity-v1} & 1911.70 $\pm$ 395.97 & 543.23 $\pm$ 316.10 & \textbf{3917.46 $\pm$ 1077.38} & 3034.83 $\pm$ 1374.72 & \textbf{4267.05 $\pm$ 678.65} & 4087.94 $\pm$ 755.10  & 4419.29 $\pm$ 232.06 & \textbf{4619.34 $\pm$ 274.43} & 3906.78 $\pm$ 795.48\\
\bottomrule
\end{tabular}
}
\end{subtable}
\caption{
The performance of \omni, which was evaluated in relation to published baselines within the \sg MuJoCo Velocity environments. Experimental outcomes, comprising mean and standard deviation, were derived from 10 assessment iterations encompassing multiple random seeds. A noteworthy distinction lies in the fact that \stable employ distinct parameters tailored to each environment, while \omni maintains a consistent parameter set across all environments. 
}
\label{compare}
\end{table}

\begin{figure}[htb]
  \centering
  \begin{subfigure}{0.49\linewidth}
    \centering
    \includegraphics[width=0.85\linewidth]{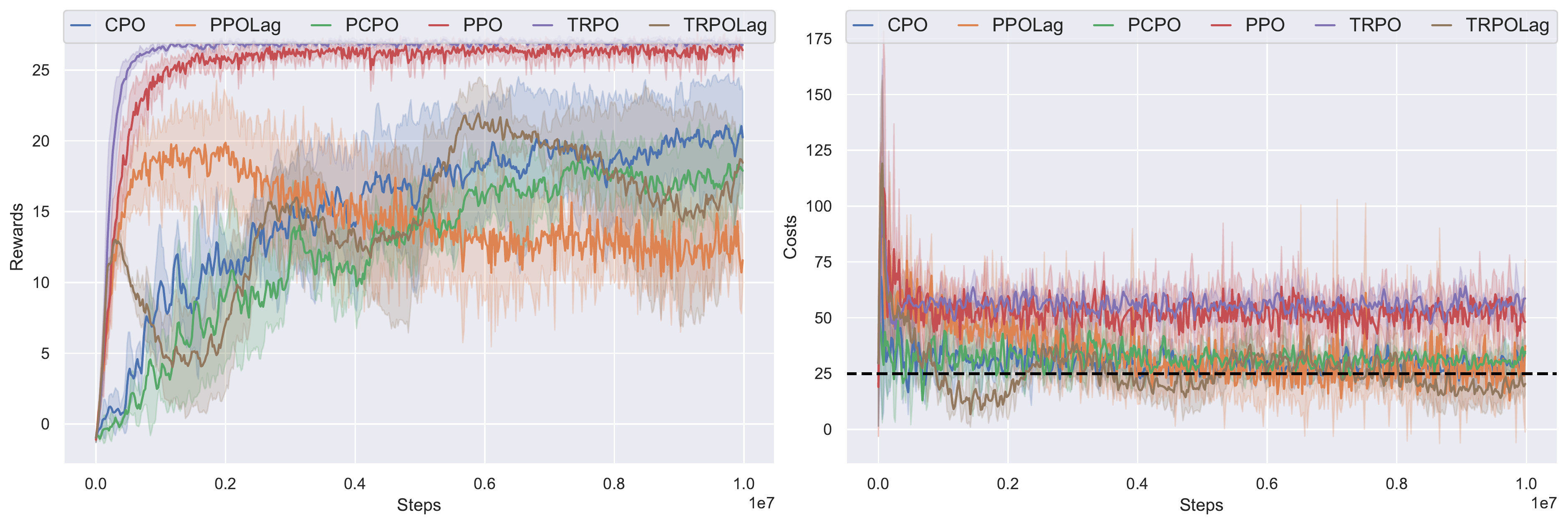}
    \caption{PointGoal1}
  \end{subfigure}
  \begin{subfigure}{0.49\linewidth}
    \centering
    \includegraphics[width=0.85\linewidth]{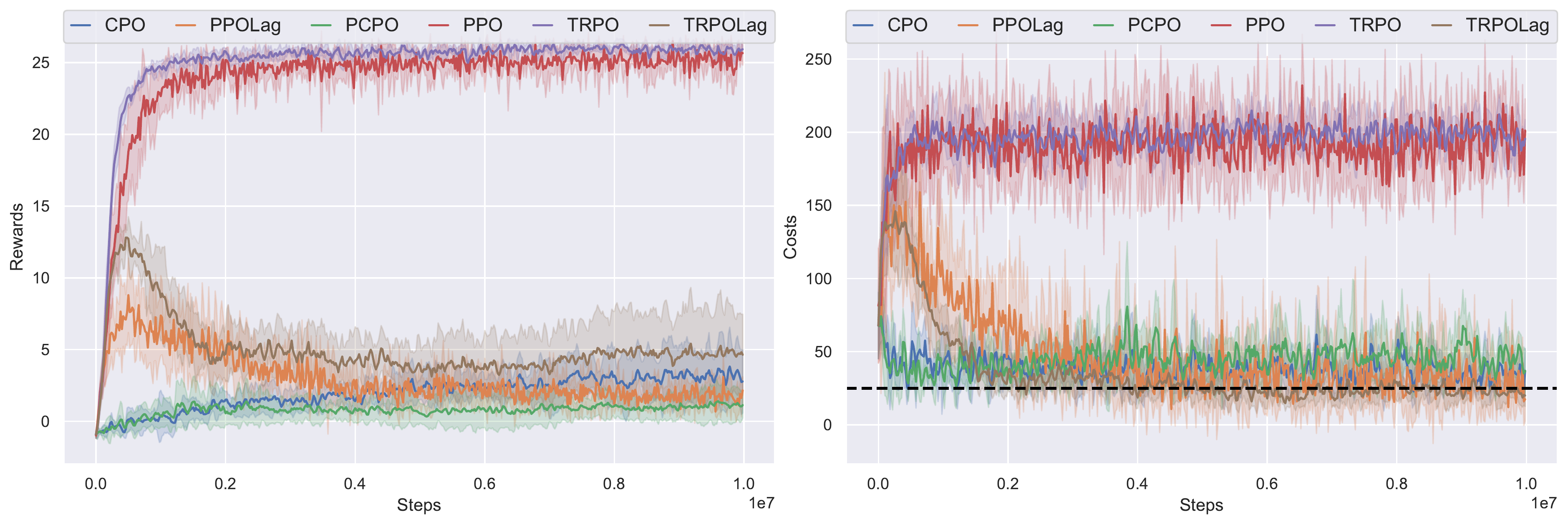}
    \caption{PointGoal2}
  \end{subfigure}
  \begin{subfigure}{0.49\linewidth}
    \centering
    \includegraphics[width=0.85\linewidth]{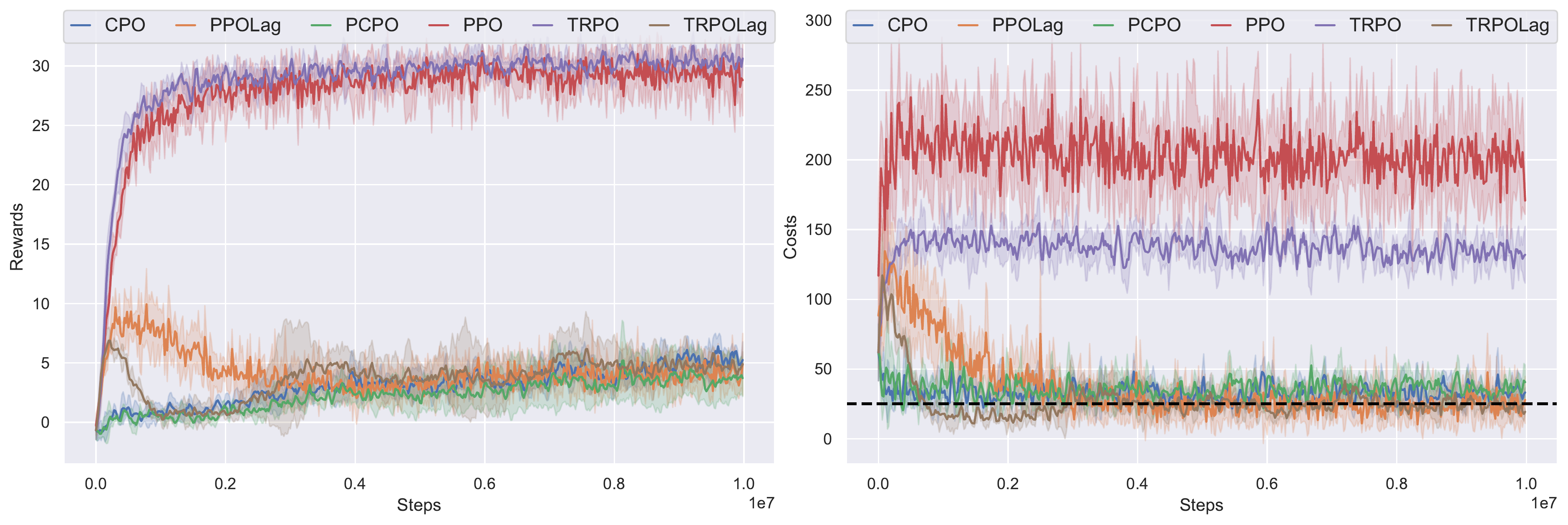}
    \caption{PointButton1}
  \end{subfigure}
  \begin{subfigure}{0.49\linewidth}
    \centering
    \includegraphics[width=0.85\linewidth]{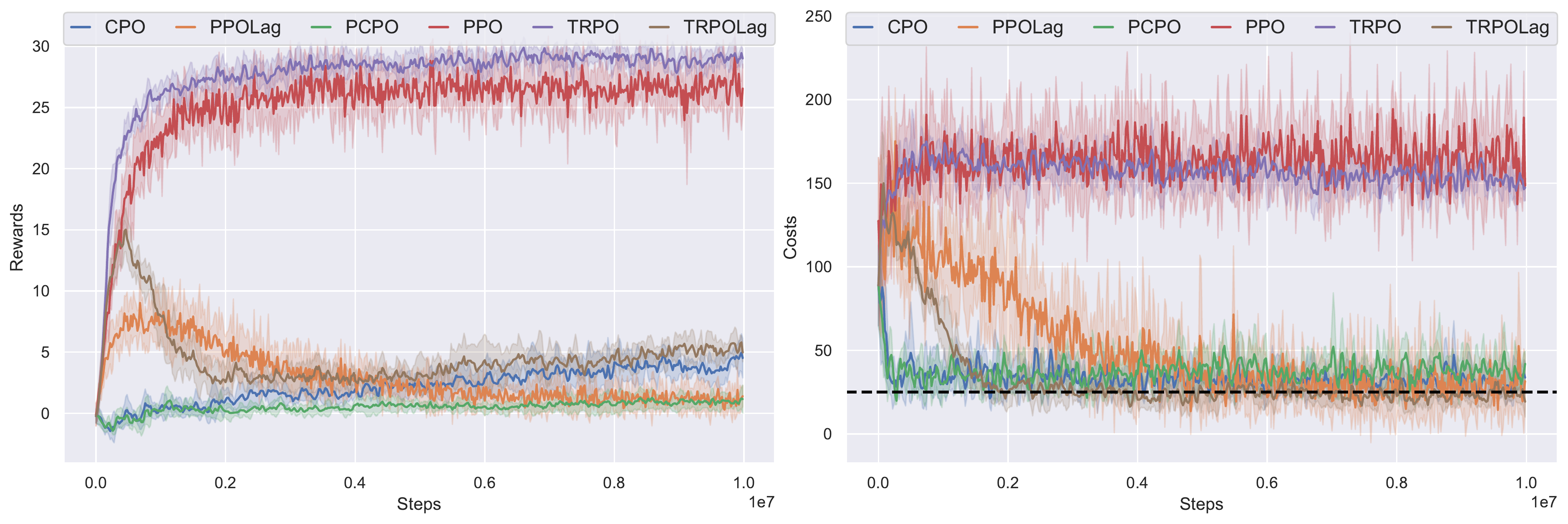}
    \caption{PointButton2}
  \end{subfigure}
  \begin{subfigure}{0.49\linewidth}
    \centering
    \includegraphics[width=0.85\linewidth]{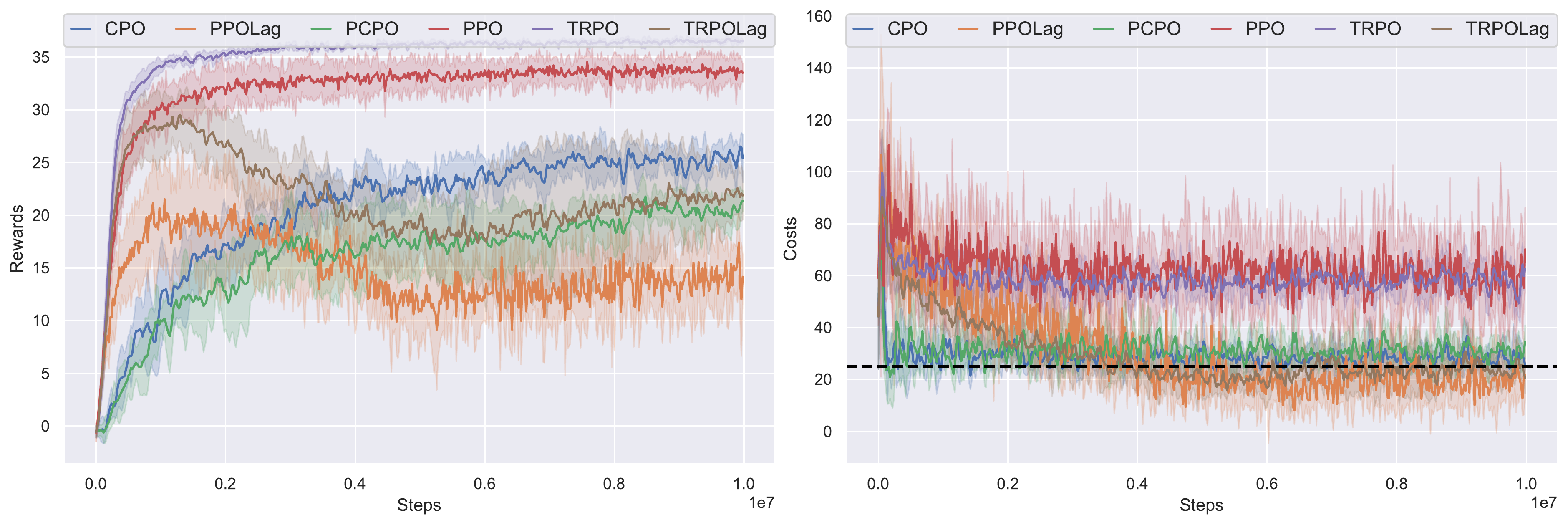}
    \caption{CarGoal1}
  \end{subfigure}
  \begin{subfigure}{0.49\linewidth}
    \centering
    \includegraphics[width=0.85\linewidth]{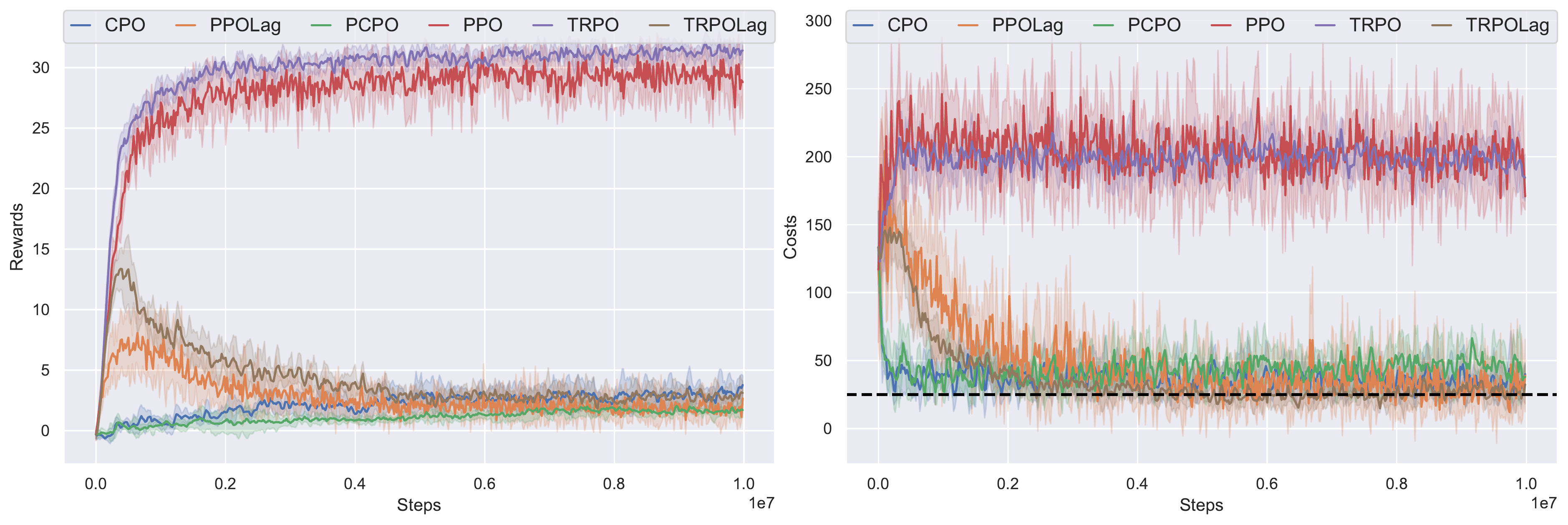}
    \caption{CarGoal2}
  \end{subfigure}
  \begin{subfigure}{0.49\linewidth}
    \centering
    \includegraphics[width=0.85\linewidth]{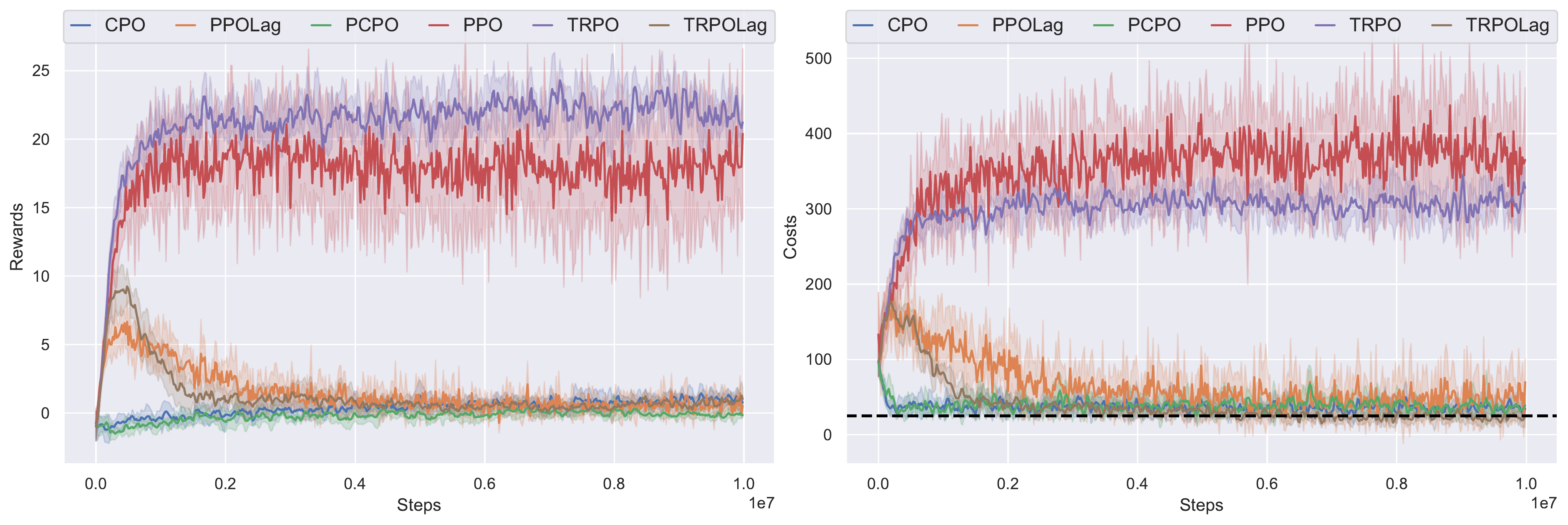}
    \caption{CarButton1}
  \end{subfigure}
  \begin{subfigure}{0.49\linewidth}
    \centering
    \includegraphics[width=0.85\linewidth]{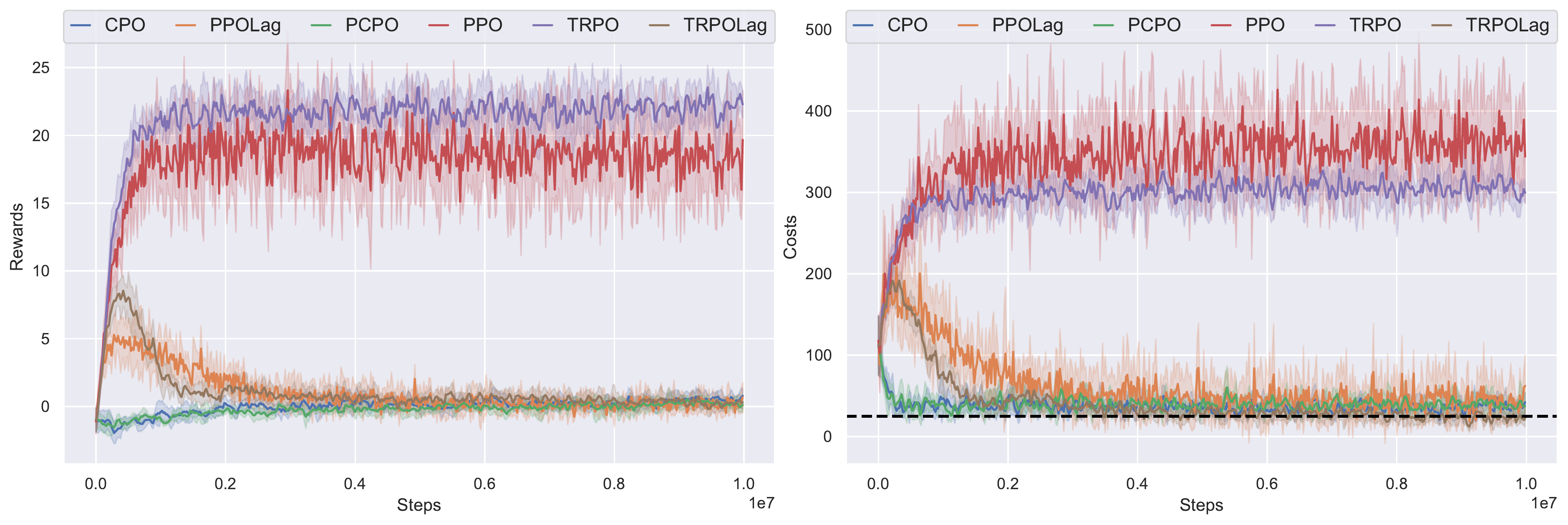}
    \caption{CarButton2}
  \end{subfigure}
  \begin{subfigure}{0.49\linewidth}
    \centering
    \includegraphics[width=0.85\linewidth]{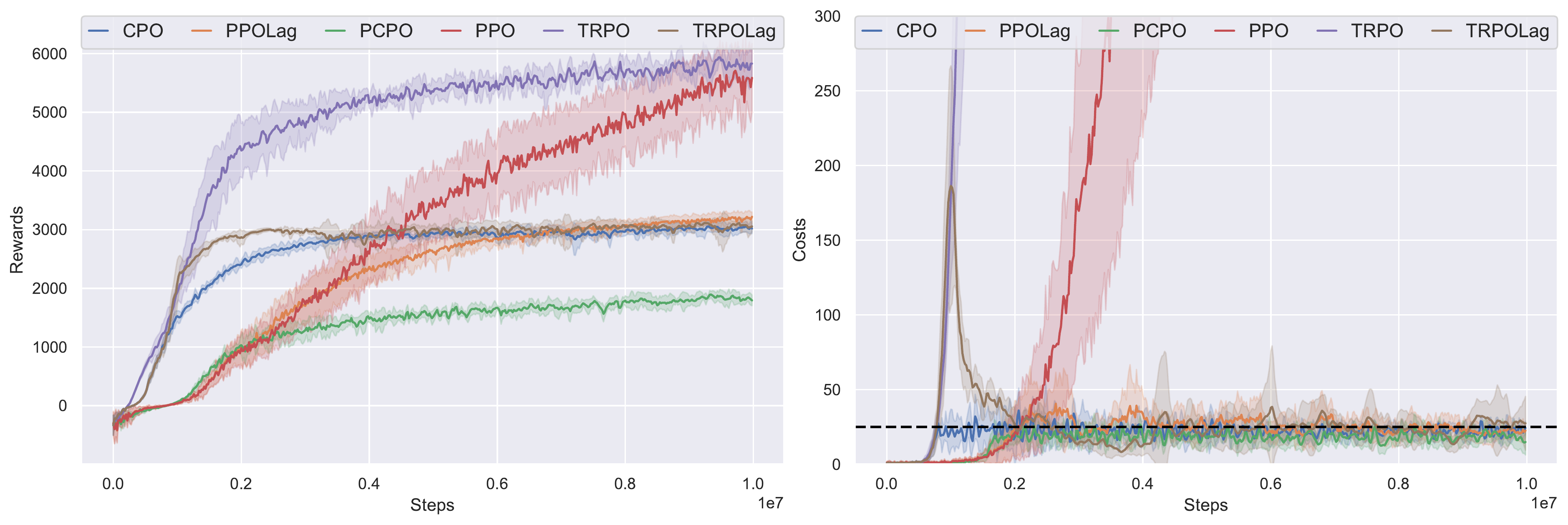}
    \caption{Ant}
  \end{subfigure}
  \begin{subfigure}{0.49\linewidth}
    \centering
    \includegraphics[width=0.85\linewidth]{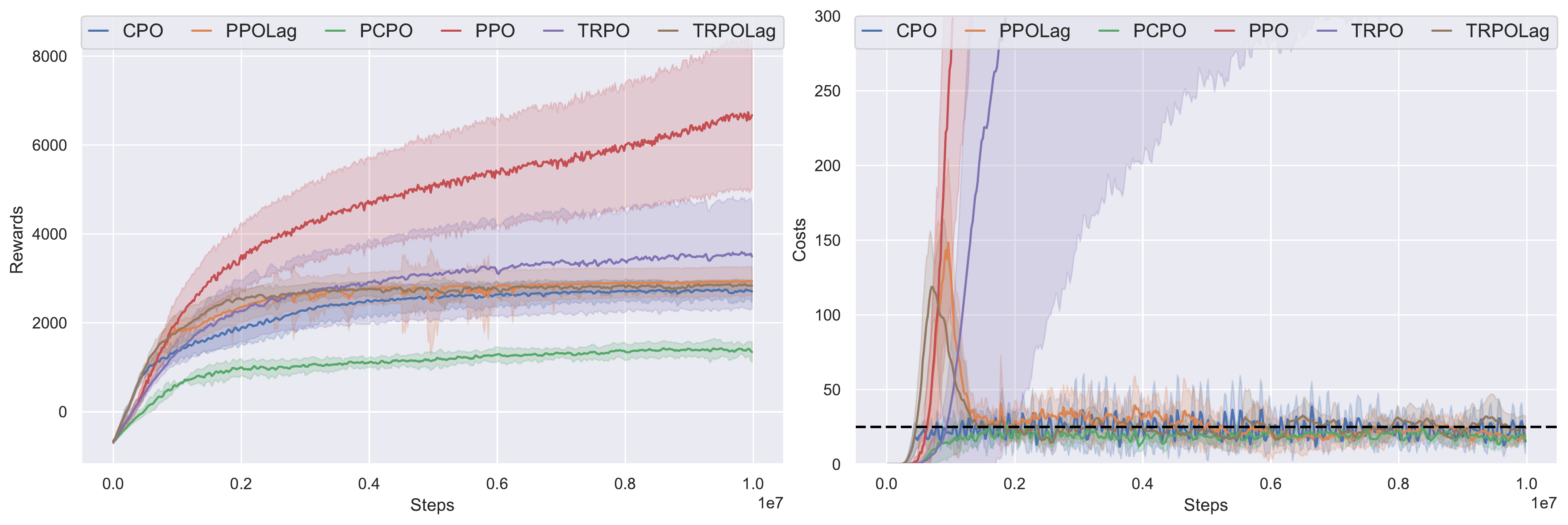}
    \caption{HalfCheetah}
  \end{subfigure}
  \begin{subfigure}{0.49\linewidth}
    \centering
    \includegraphics[width=0.85\linewidth]{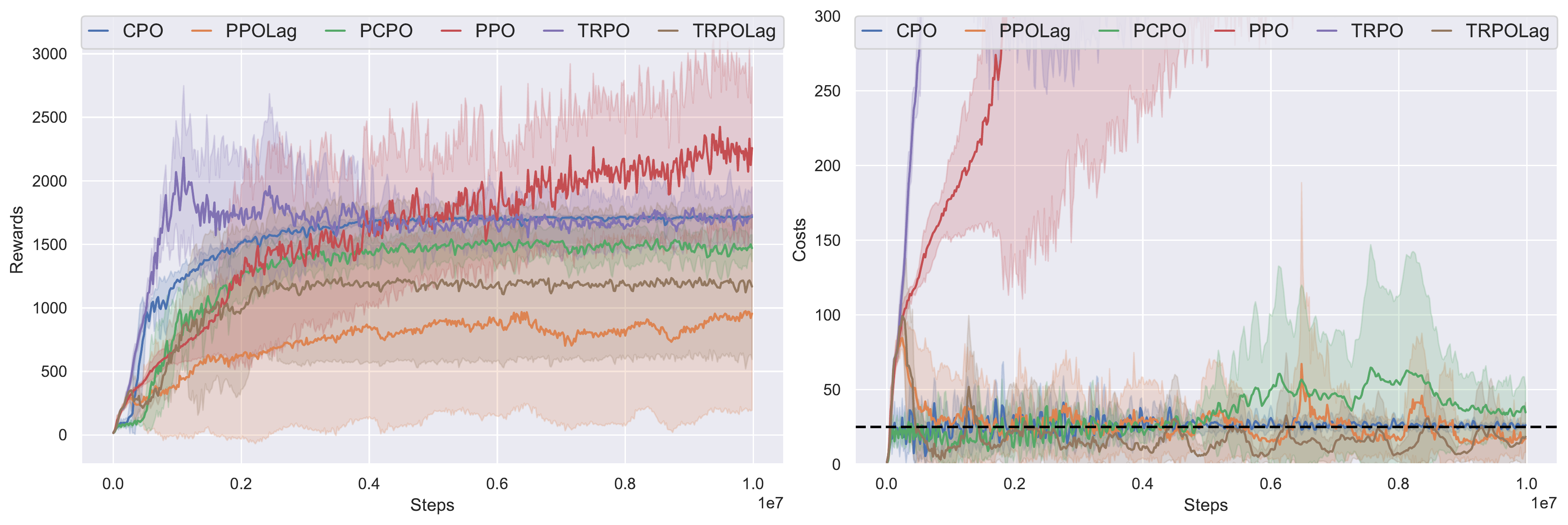}
    \caption{Hopper}
  \end{subfigure}
  \begin{subfigure}{0.49\linewidth}
    \centering
    \includegraphics[width=0.85\linewidth]{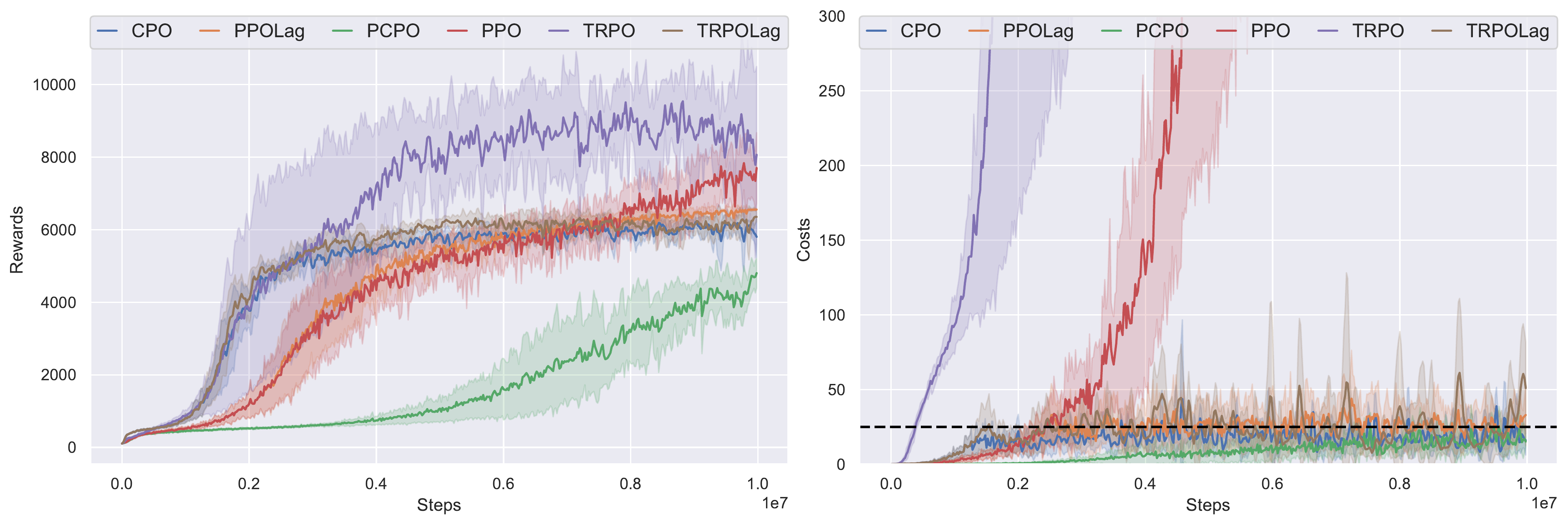}
    \caption{Humanoid}
  \end{subfigure}
  \begin{subfigure}{0.49\linewidth}
    \centering
    \includegraphics[width=0.85\linewidth]{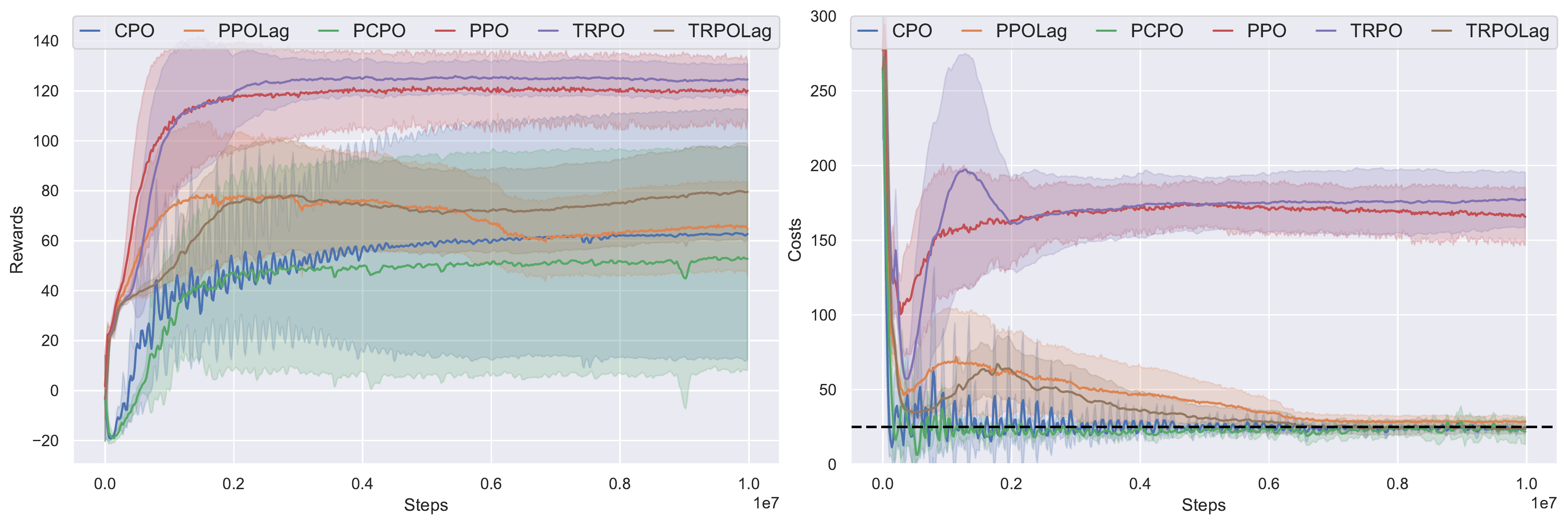}
    \caption{Swimmer}
  \end{subfigure}
  \begin{subfigure}{0.49\linewidth}
    \centering
    \includegraphics[width=0.85\linewidth]{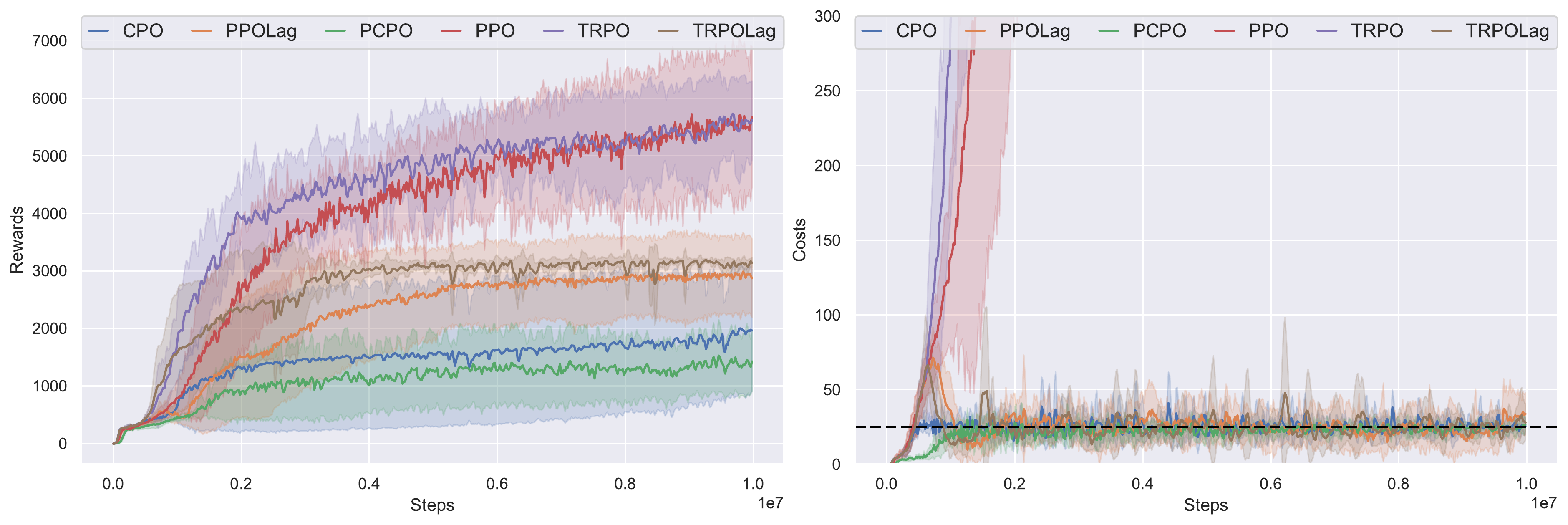}
    \caption{Walker2d}
  \end{subfigure}
\caption{Training curves in \sg MuJoCo Navigation and Velocity environments. The rewards and costs are obtained from 1e7 steps interaction. Dashed black lines indicate the target cost value for a safe policy, which is set as 25.0.}
\label{benchmark}
\end{figure}

\noindent
\omni provides experiment results conducted in the \sg environment, as illustrated partially in \autoref{benchmark}. For a comprehensive summary of \omni algorithm performance\footnote{The rule for Early-Terminated MDP is to stop when the constraint violation value exceeds the set range. Therefore, for some related algorithms in \autoref{performance}, their cost values happen to be exactly 25.00.}, please refer to \autoref{performance}.

\clearpage

\begin{table}[htb]

\captionsetup[subtable]{justification=centering}

\begin{subtable}{\linewidth}\centering
\resizebox{\columnwidth}{!}{
\begin{tabular}{@{}l|cc|cc|cc|cc@{}}\toprule
    & \multicolumn{2}{c|}{\textbf{Policy Gradient}} & \multicolumn{2}{c|}{\textbf{Natural PG}} & \multicolumn{2}{c|}{\textbf{TRPO}} & \multicolumn{2}{c}{\textbf{PPO}} \\
\midrule
\textbf{Environment} \hfill  & \textbf{Reward} & \textbf{Cost} & \textbf{Reward}  & \textbf{Cost}  & \textbf{Reward} & \textbf{Cost} & \textbf{Reward} & \textbf{Cost}  \\ \midrule
\textsc{SafetyAntVelocity-v1} & 5292.29 $\pm$ 913.44 & 919.42 $\pm$ 158.61 & 5547.20 $\pm$ 807.89 & 895.56 $\pm$ 77.13 & 6026.79 $\pm$ 314.98  & 933.46 $\pm$ 41.28 & 5977.73 $\pm$ 885.65 & 958.13 $\pm$ 134.50  \\
\textsc{SafetyHalfCheetahVelocity-v1} & 5188.46 $\pm$ 1202.76 & 896.55 $\pm$ 184.70 & 5878.28 $\pm$ 2012.24 & 847.74 $\pm$ 249.02 & 6490.76 $\pm$ 2507.18  & 734.26 $\pm$ 321.88 & 6921.83 $\pm$ 1721.79 & 919.20 $\pm$ 173.08  \\
\textsc{SafetyHopperVelocity-v1} & 3218.17 $\pm$ 672.88 & 881.76 $\pm$ 198.46 & 2613.95 $\pm$ 866.13 & 587.78 $\pm$ 220.97 & 2047.35 $\pm$ 447.33  & 448.12 $\pm$ 103.87 & 2337.11 $\pm$ 942.06 & 550.02 $\pm$ 237.70   \\
\textsc{SafetyHumanoidVelocity-v1} & 7001.78 $\pm$ 419.67 & 834.11 $\pm$ 212.43 & 8055.20 $\pm$ 641.67 & 946.40 $\pm$ 9.11 & 8681.24 $\pm$ 3934.08  & 718.42 $\pm$ 323.30 & 9115.93 $\pm$ 596.88 & 960.44 $\pm$ 7.06   \\
\textsc{SafetySwimmerVelocity-v1} & 77.05 $\pm$ 33.44 & 107.10 $\pm$ 60.58 & 120.19 $\pm$ 7.74 & 161.78 $\pm$ 17.51 & 124.91 $\pm$ 6.13  & 176.56 $\pm$ 15.95 & 119.77 $\pm$ 13.80 & 165.27 $\pm$ 20.15   \\
\textsc{SafetyWalker2dVelocity-v1} & 4832.34 $\pm$ 685.76 & 866.59 $\pm$ 93.47 & 5347.35 $\pm$ 436.86 & 914.74 $\pm$ 32.61 & 6096.67 $\pm$ 723.06  & 914.46 $\pm$ 27.85 & 6239.52 $\pm$ 879.99 & 902.68 $\pm$ 100.93  \\
\textsc{SafetyCarGoal1-v0} & 35.86 $\pm$ 1.97 & 57.46 $\pm$ 48.34 & 36.07 $\pm$ 1.25 & 58.06 $\pm$ 10.03 & 36.60 $\pm$ 0.22  & 55.58 $\pm$ 12.68 & 33.41 $\pm$ 2.89 & 58.06 $\pm$ 42.06   \\
\textsc{SafetyCarButton1-v0} & 19.76 $\pm$ 10.15 & 353.26 $\pm$ 177.08 & 22.16 $\pm$ 4.48 & 333.98 $\pm$ 67.49 & 21.98 $\pm$ 2.06  & 343.22 $\pm$ 24.60 & 17.51 $\pm$ 9.46 & 373.98 $\pm$ 156.64   \\
\textsc{SafetyCarGoal2-v0} & 29.43 $\pm$ 4.62 & 179.20 $\pm$ 84.86  & 30.26 $\pm$ 0.38 & 209.62 $\pm$ 29.97 & 32.17 $\pm$ 1.24  & 190.74 $\pm$ 21.05 & 29.88 $\pm$ 4.55 & 194.16 $\pm$ 106.20   \\
\textsc{SafetyCarButton2-v0} & 18.06 $\pm$ 10.53 & 349.82 $\pm$ 187.07 & 20.85 $\pm$ 3.14 & 313.88 $\pm$ 58.20 & 20.51 $\pm$ 3.34  & 316.42 $\pm$ 35.28 & 21.35 $\pm$ 8.22 & 312.64 $\pm$ 138.40   \\
\textsc{SafetyPointGoal1-v0} & 26.19 $\pm$ 3.44 & 201.22 $\pm$ 80.40 & 26.92 $\pm$ 0.58 & 57.92 $\pm$ 9.97 & 27.20 $\pm$ 0.44   & 45.88 $\pm$ 11.27 & 25.44 $\pm$ 5.43 & 55.72 $\pm$ 35.55   \\
\textsc{SafetyPointButton1-v0} & 29.98 $\pm$ 5.24 & 141.74 $\pm$ 75.13 & 31.95 $\pm$ 1.53 & 123.98 $\pm$ 32.05 & 30.61 $\pm$ 0.40  & 134.38 $\pm$ 22.06 & 27.03 $\pm$ 6.14 & 152.48 $\pm$ 80.39   \\
\textsc{SafetyPointGoal2-v0} & 25.18 $\pm$ 3.62 & 204.96 $\pm$ 104.97 & 26.19 $\pm$ 0.84 & 193.60 $\pm$ 18.54 & 25.61 $\pm$ 0.89  & 202.26 $\pm$ 15.15 & 25.49 $\pm$ 2.46 & 159.28 $\pm$ 87.13   \\
\textsc{SafetyPointButton2-v0} & 26.88 $\pm$ 4.38 & 153.88 $\pm$ 65.54 & 28.45 $\pm$ 1.49 & 160.40 $\pm$ 20.08 & 28.78 $\pm$ 2.05  & 170.30 $\pm$ 30.59 & 25.91 $\pm$ 6.15 & 166.60 $\pm$ 111.21   \\
\bottomrule
\end{tabular}
}
\caption{On-Policy Base Algorithms. \\ \hspace*{\fill} \\}
\label{on-base}
\end{subtable}

\begin{subtable}{\linewidth}\centering
\resizebox{\columnwidth}{!}{
\begin{tabular}{@{}l|cc|cc|cc|cc@{}}
\toprule
    & \multicolumn{2}{c|}{\textbf{RCPO}} & \multicolumn{2}{c|}{\textbf{TRPOLag}} & \multicolumn{2}{c|}{\textbf{PPOLag}}  & \multicolumn{2}{c}{\textbf{P3O}} \\
\midrule
\textbf{Environment} \hfill & \textbf{Reward}  & \textbf{Cost}  & \textbf{Reward} & \textbf{Cost} & \textbf{Reward} & \textbf{Cost} & \textbf{Reward} & \textbf{Cost}  \\ \midrule
\textsc{SafetyAntVelocity-v1} & 3139.52 $\pm$ 110.34 & 12.34 $\pm$ 3.11  & 3041.89 $\pm$ 180.77  & 19.52 $\pm$ 20.21 & 3261.87 $\pm$ 80.00 & 12.05 $\pm$ 6.57  & 2636.62 $\pm$ 181.09 & 20.69 $\pm$ 10.23 \\
\textsc{SafetyHalfCheetahVelocity-v1} & 2440.97 $\pm$ 451.88 & 9.02 $\pm$ 9.34 & 2884.68 $\pm$ 77.47  & 9.04 $\pm$ 11.83 & 2946.15 $\pm$ 306.35 & 3.44 $\pm$ 4.77 & 2117.84 $\pm$ 313.55 & 27.60 $\pm$ 8.36  \\
\textsc{SafetyHopperVelocity-v1} & 1428.58 $\pm$ 199.87 & 11.12 $\pm$ 12.66 & 1391.79 $\pm$ 269.07  & 11.22 $\pm$ 9.97 & 961.92 $\pm$ 752.87 & 13.96 $\pm$ 19.33 & 1231.52 $\pm$ 465.35 & 16.33 $\pm$ 11.38  \\
\textsc{SafetyHumanoidVelocity-v1}  & 6286.51 $\pm$ 151.03 & 19.47 $\pm$ 7.74 & 6551.30 $\pm$ 58.42  & 59.56 $\pm$ 117.37 & 6624.46 $\pm$ 25.90 & 5.87 $\pm$ 9.46 & 6342.47 $\pm$ 82.45 & 126.40 $\pm$ 193.76 \\
\textsc{SafetySwimmerVelocity-v1}  & 61.29 $\pm$ 18.12 & 22.60 $\pm$ 1.16 & 81.18 $\pm$ 16.33  & 22.24 $\pm$ 3.91 & 64.74 $\pm$ 17.67 & 28.02 $\pm$ 4.09 & 38.02 $\pm$ 34.18 & 18.40 $\pm$ 12.13  \\
\textsc{SafetyWalker2dVelocity-v1}  & 3064.43 $\pm$ 218.83 & 3.02 $\pm$ 1.48 & 3207.10 $\pm$ 7.88  & 14.98 $\pm$ 9.27 & 2982.27 $\pm$ 681.55 & 13.49 $\pm$ 14.55 & 2713.57 $\pm$ 313.20 & 20.51 $\pm$ 14.09  \\
\textsc{SafetyCarGoal1-v0} & 18.71 $\pm$ 2.72 & 23.10 $\pm$ 12.57 & 27.04 $\pm$ 1.82   & 26.80 $\pm$ 5.64 & 13.27 $\pm$ 9.26 & 21.72 $\pm$ 32.06  & -1.10 $\pm$ 6.851 & 50.58 $\pm$ 99.24 \\
\textsc{SafetyCarButton1-v0} & -2.04 $\pm$ 2.98 & 43.48 $\pm$ 31.52 & -0.38 $\pm$ 0.85   & 37.54 $\pm$ 31.72 & 0.33 $\pm$ 1.96 & 55.50 $\pm$ 89.64 & -2.06 $\pm$ 7.20 & 43.78 $\pm$ 98.01  \\
\textsc{SafetyCarGoal2-v0} & 2.30 $\pm$ 1.76 & 22.90 $\pm$ 16.22 & 3.65 $\pm$ 1.09   & 39.98 $\pm$ 20.29 & 1.58 $\pm$ 2.49 & 13.82 $\pm$ 24.62 & -0.07 $\pm$ 1.62 & 43.86 $\pm$ 99.58  \\
\textsc{SafetyCarButton2-v0} & -1.35 $\pm$ 2.41 & 42.02 $\pm$ 31.77 & -1.68 $\pm$ 2.55   & 20.36 $\pm$ 13.67 & 0.76 $\pm$ 2.52 & 47.86 $\pm$ 103.27 & 0.11 $\pm$ 0.72 & 85.94 $\pm$ 122.01  \\
\textsc{SafetyPointGoal1-v0} & 15.27 $\pm$ 4.05 & 30.56 $\pm$ 19.15 & 18.51 $\pm$ 3.83   & 22.98 $\pm$ 8.45 & 12.96 $\pm$ 6.95 & 25.80 $\pm$ 34.99 & 1.60 $\pm$ 3.01 & 31.10 $\pm$ 80.03  \\
\textsc{SafetyPointButton1-v0} & 3.65 $\pm$ 4.47 & 26.30 $\pm$ 9.22 & 6.93 $\pm$ 1.84   & 31.16 $\pm$ 20.58 & 4.60 $\pm$ 4.73 & 20.80 $\pm$ 35.78 & -0.34 $\pm$ 1.53 & 52.86 $\pm$ 85.62  \\
\textsc{SafetyPointGoal2-v0} & 2.17 $\pm$ 1.46 & 33.82 $\pm$ 21.93 & 4.64 $\pm$ 1.43   & 26.00 $\pm$ 4.70 & 1.98 $\pm$ 3.86 & 41.20 $\pm$ 61.03 & 0.34 $\pm$ 2.20 & 65.84 $\pm$ 195.76  \\
\textsc{SafetyPointButton2-v0} & 7.18 $\pm$ 1.93 & 45.02 $\pm$ 25.28 & 5.43 $\pm$ 3.44   & 25.10 $\pm$ 8.98 & 0.93 $\pm$ 3.69 & 33.72 $\pm$ 58.75 & 0.33 $\pm$ 2.44 & 28.50 $\pm$ 49.79  \\
\bottomrule
\end{tabular}
}
\end{subtable}

\begin{subtable}{\linewidth}\centering
\resizebox{\columnwidth}{!}{
\begin{tabular}{@{}l|cc|cc|cc|cc@{}}\toprule
    & \multicolumn{2}{c|}{\textbf{CUP}} & \multicolumn{2}{c|}{\textbf{PCPO}} & \multicolumn{2}{c|}{\textbf{FOCOPS}} & \multicolumn{2}{c}{\textbf{CPO}} \\
\midrule
\textbf{Environment} \hfill  & \textbf{Reward} & \textbf{Cost} & \textbf{Reward}  & \textbf{Cost}  & \textbf{Reward} & \textbf{Cost} & \textbf{Reward} & \textbf{Cost}  \\ \midrule
\textsc{SafetyAntVelocity-v1} & 3215.79 $\pm$ 346.68 & 18.25 $\pm$ 17.12 & 2257.07 $\pm$ 47.97 & 10.44 $\pm$ 5.22 & 3184.48 $\pm$ 305.59  & 14.75 $\pm$ 6.36 & 3098.54 $\pm$ 78.90 & 14.12 $\pm$ 3.41   \\
\textsc{SafetyHalfCheetahVelocity-v1} & 2850.60 $\pm$ 244.65 & 4.27 $\pm$ 4.46 & 1677.93 $\pm$ 217.31 & 19.06 $\pm$ 15.26 & 2965.20 $\pm$ 290.43  &  2.37 $\pm$ 3.50 & 2786.48 $\pm$ 173.45 & 4.70 $\pm$ 6.72   \\
\textsc{SafetyHopperVelocity-v1} & 1716.08 $\pm$ 5.93 & 7.48 $\pm$ 5.535 & 1551.22 $\pm$ 85.16 & 15.46 $\pm$ 9.83 & 1437.75 $\pm$ 446.87  & 10.13 $\pm$ 8.87 & 1713.71 $\pm$ 18.26 & 13.40 $\pm$ 5.82   \\
\textsc{SafetyHumanoidVelocity-v1} & 6109.94 $\pm$ 497.56 & 24.69 $\pm$ 20.54 & 5852.25 $\pm$ 78.01 & 0.24 $\pm$ 0.48 & 6489.39 $\pm$ 35.10  & 13.86 $\pm$ 39.33 & 6465.34 $\pm$ 79.87 & 0.18 $\pm$ 0.36   \\
\textsc{SafetySwimmerVelocity-v1} & 63.83 $\pm$ 46.45 & 21.95 $\pm$ 11.04 & 54.42 $\pm$ 38.65 & 17.34 $\pm$ 1.57 & 53.87 $\pm$ 17.90  & 29.75 $\pm$ 7.33 & 65.30 $\pm$ 43.25 & 18.22 $\pm$ 8.01   \\
\textsc{SafetyWalker2dVelocity-v1} & 2466.95 $\pm$ 1114.13 & 6.63 $\pm$ 8.25 & 1802.86 $\pm$ 714.04 & 18.82 $\pm$ 5.57 & 3117.05 $\pm$ 53.60  & 8.78 $\pm$ 12.38 & 2074.76 $\pm$ 962.45 & 21.90 $\pm$ 9.41   \\
\textsc{SafetyCarGoal1-v0} & 6.14 $\pm$ 6.97 & 36.12 $\pm$ 89.56 & 21.56 $\pm$ 2.87 & 38.42 $\pm$ 8.36 & 15.23 $\pm$ 10.76  & 31.66 $\pm$ 93.51 & 25.52 $\pm$ 2.65 & 43.32 $\pm$ 14.35   \\
\textsc{SafetyCarButton1-v0} & 1.49 $\pm$ 2.84 & 103.24 $\pm$ 123.12 & 0.36 $\pm$ 0.85 & 40.52 $\pm$ 21.25 & 0.21 $\pm$ 2.27  & 31.78 $\pm$ 47.03 & 0.82 $\pm$ 1.60 & 37.86 $\pm$ 27.41   \\
\textsc{SafetyCarGoal2-v0} & 1.78 $\pm$ 4.03 & 95.40 $\pm$ 129.64  & 1.62 $\pm$ 0.56 & 48.12 $\pm$ 31.19 & 2.09 $\pm$ 4.33  & 31.56 $\pm$ 58.93 & 3.56 $\pm$ 0.92 & 32.66 $\pm$ 3.31   \\
\textsc{SafetyCarButton2-v0} & 1.49 $\pm$ 2.64 & 173.68 $\pm$ 163.77 & 0.66 $\pm$ 0.42 & 49.72 $\pm$ 36.50 & 1.14 $\pm$ 3.18  & 46.78 $\pm$ 57.47 & 0.17 $\pm$ 1.19 & 48.56 $\pm$ 29.34   \\
\textsc{SafetyPointGoal1-v0} & 14.42 $\pm$ 6.74 & 19.02 $\pm$ 20.08 & 18.57 $\pm$ 1.71 & 22.98 $\pm$ 6.56 & 14.97 $\pm$ 9.01   & 33.72 $\pm$ 42.24 & 20.46 $\pm$ 1.38 & 28.84 $\pm$ 7.76   \\
\textsc{SafetyPointButton1-v0} & 3.50 $\pm$ 7.07 & 39.56 $\pm$ 54.26  & 2.66 $\pm$ 1.83 & 49.40 $\pm$ 36.76 & 5.89 $\pm$ 7.66  & 38.24 $\pm$ 42.96 & 4.04 $\pm$ 4.54 & 40.00 $\pm$ 4.52   \\
\textsc{SafetyPointGoal2-v0} & 1.06 $\pm$ 2.67 & 107.30 $\pm$ 204.26 & 1.06 $\pm$ 0.69 & 51.92 $\pm$ 47.40 & 2.21 $\pm$ 4.15  & 37.92 $\pm$ 111.81 & 2.50 $\pm$ 1.25 & 40.84 $\pm$ 23.31   \\
\textsc{SafetyPointButton2-v0} & 2.88 $\pm$ 3.65 & 54.24 $\pm$ 71.07 & 1.05 $\pm$ 1.27 & 41.14 $\pm$ 12.35 & 2.43 $\pm$ 3.33  & 17.92 $\pm$ 26.10 & 5.09 $\pm$ 1.83 & 48.92 $\pm$ 17.79   \\
\bottomrule
\end{tabular}
}
\end{subtable}

\begin{subtable}{\linewidth}\centering
\resizebox{\columnwidth}{!}{
\begin{tabular}{@{}l|cc|cc|cc|cc@{}}\toprule
    & \multicolumn{2}{c|}{\textbf{PPOSaute}} & \multicolumn{2}{c|}{\textbf{TRPOSaute}} & \multicolumn{2}{c|}{\textbf{PPOSimmerPID}} & \multicolumn{2}{c}{\textbf{TRPOSimmerPID}} \\
\midrule
\textbf{Environment} \hfill  & \textbf{Reward} & \textbf{Cost} & \textbf{Reward}  & \textbf{Cost}  & \textbf{Reward} & \textbf{Cost} & \textbf{Reward} & \textbf{Cost}  \\ \midrule
\textsc{SafetyAntVelocity-v1} & 2978.74 $\pm$ 93.65 & 16.77 $\pm$ 0.92 & 2507.65 $\pm$ 63.97 & 8.036 $\pm$ 0.39 & 2944.84 $\pm$ 60.53  & 16.20 $\pm$ 0.66 & 3018.95 $\pm$ 66.44 & 16.52 $\pm$ 0.23   \\
\textsc{SafetyHalfCheetahVelocity-v1} & 2901.40 $\pm$ 25.49 & 16.20 $\pm$ 0.60 & 2521.80 $\pm$ 477.29 & 7.61 $\pm$ 0.39 & 2922.17 $\pm$ 24.84 & 16.14 $\pm$ 0.14 & 2737.79 $\pm$ 37.53 & 16.44 $\pm$ 0.21   \\
\textsc{SafetyHopperVelocity-v1} & 1650.91 $\pm$ 152.65 & 17.87 $\pm$ 1.33 & 1368.28 $\pm$ 576.08 & 10.38 $\pm$ 4.38 & 1699.94 $\pm$ 24.25  & 17.04 $\pm$ 0.41 & 1608.41 $\pm$ 88.23 & 16.30 $\pm$ 0.30   \\
\textsc{SafetyHumanoidVelocity-v1} & 6401.00 $\pm$ 32.23 & 17.10 $\pm$ 2.41 & 5759.44 $\pm$ 75.73 & 15.84 $\pm$ 1.42 & 6401.85 $\pm$ 57.62  & 11.06 $\pm$ 5.35 & 6411.32 $\pm$ 44.26 & 13.04 $\pm$ 2.68   \\
\textsc{SafetySwimmerVelocity-v1} & 35.61 $\pm$ 4.37 & 3.44 $\pm$ 1.35 & 34.72 $\pm$ 1.37 & 10.19 $\pm$ 2.32 & 77.52 $\pm$ 40.20  & 0.98 $\pm$ 1.91 & 51.39 $\pm$ 40.09 & 0.00 $\pm$ 0.00   \\
\textsc{SafetyWalker2dVelocity-v1} & 2410.89 $\pm$ 241.22 & 18.88 $\pm$ 2.38 & 2548.82 $\pm$ 891.65 & 13.21 $\pm$ 6.09 & 3187.56 $\pm$ 32.66  & 17.10 $\pm$ 0.49 & 3156.99 $\pm$ 30.93 & 17.14 $\pm$ 0.54   \\
\textsc{SafetyCarGoal1-v0} & -0.65 $\pm$ 2.89 & 22.90 $\pm$ 16.85 & 1.89 $\pm$ 3.52 & 4.86 $\pm$ 3.11 & 0.81 $\pm$ 0.41  & 17.18 $\pm$ 14.30 & 0.24 $\pm$ 0.41 & 0.90 $\pm$ 1.18   \\
\textsc{SafetyCarButton1-v0} & -1.72 $\pm$ 0.89 & 51.88 $\pm$ 28.18 & -2.03 $\pm$ 0.40 & 6.24 $\pm$ 6.14 & -0.57 $\pm$ 0.63  & 49.14 $\pm$ 37.77 & -1.24 $\pm$ 0.47 & 17.26 $\pm$ 16.13   \\
\textsc{SafetyCarGoal2-v0} & -0.87 $\pm$ 0.79 & 6.13 $\pm$ 4.51  & -1.03 $\pm$ 1.46 & 18.07 $\pm$ 11.62 & -0.96 $\pm$ 1.10  & 3.00 $\pm$ 0.83 & -0.36 $\pm$ 0.09 & 9.83 $\pm$ 13.91  \\
\textsc{SafetyCarButton2-v0} & -1.89 $\pm$ 1.86 & 47.33 $\pm$ 28.90 & -2.60 $\pm$ 0.40 & 74.57 $\pm$ 84.95 & -1.31 $\pm$ 0.93  & 52.33 $\pm$ 19.96 & -0.99 $\pm$ 0.63 & 20.40 $\pm$ 12.77   \\
\textsc{SafetyPointGoal1-v0} & 1.99 $\pm$ 2.87 & 7.80 $\pm$ 2.78 & 1.02 $\pm$ 0.80 & 7.46 $\pm$ 5.26 & 1.69 $\pm$ 3.25   & 5.34 $\pm$ 10.33 & 1.38 $\pm$ 1.91 & 1.34 $\pm$ 1.82   \\
\textsc{SafetyPointButton1-v0} & -1.47 $\pm$ 0.98 & 22.60 $\pm$ 13.91 & -3.13 $\pm$ 3.51 & 9.04 $\pm$ 3.94 & -1.97 $\pm$ 1.41  & 12.80 $\pm$ 7.84 & -1.36 $\pm$ 0.37 & 2.14 $\pm$ 1.73   \\
\textsc{SafetyPointGoal2-v0} & -1.85 $\pm$ 0.99 & 21.77 $\pm$ 13.56 & -1.38 $\pm$ 1.16 & 7.87 $\pm$ 2.02 & -1.13 $\pm$ 0.39  & 7.03 $\pm$ 4.21 & -0.54 $\pm$ 0.18 & 26.57 $\pm$ 19.13   \\
\textsc{SafetyPointButton2-v0} & -1.38 $\pm$ 0.11 & 12.00 $\pm$ 8.60 & -2.56 $\pm$ 0.67 & 17.27 $\pm$ 10.01 & -1.70 $\pm$ 0.29  & 7.90 $\pm$ 3.30 & -1.66 $\pm$ 0.99 & 6.70 $\pm$ 4.74   \\
\bottomrule
\end{tabular}
}
\end{subtable}

\begin{subtable}{\linewidth}\centering
\resizebox{\columnwidth}{!}{
\begin{tabular}{@{}l|cc|cc|cc|cc@{}}\toprule
    & \multicolumn{2}{c|}{\textbf{CPPOPID}} & \multicolumn{2}{c|}{\textbf{TRPOPID}} & \multicolumn{2}{c|}{\textbf{PPOEarlyTerminated}} & \multicolumn{2}{c}{\textbf{TRPOEarlyTerminated}} \\
\midrule
\textbf{Environment} \hfill  & \textbf{Reward} & \textbf{Cost} & \textbf{Reward}  & \textbf{Cost}  & \textbf{Reward} & \textbf{Cost} & \textbf{Reward} & \textbf{Cost}  \\ \midrule
\textsc{SafetyAntVelocity-v1} & 3213.36 $\pm$ 146.78 & 14.30 $\pm$ 7.39 & 3052.94 $\pm$ 139.67 & 15.22 $\pm$ 3.68 & 2801.53 $\pm$ 19.66  & 0.23 $\pm$ 0.09 & 3052.63 $\pm$ 58.41 & 0.40 $\pm$ 0.23   \\
\textsc{SafetyHalfCheetahVelocity-v1} & 2837.89 $\pm$ 398.52 & 8.06 $\pm$ 9.62 & 2796.75 $\pm$ 190.84 & 11.16 $\pm$ 9.80 & 2447.25 $\pm$ 346.84  & 3.47 $\pm$ 4.90 & 2555.70 $\pm$ 368.17 & 0.06 $\pm$ 0.08   \\
\textsc{SafetyHopperVelocity-v1} & 1713.29 $\pm$ 10.21 & 8.96 $\pm$ 4.28 & 1178.59 $\pm$ 646.71 & 18.76 $\pm$ 8.93 & 1643.39 $\pm$ 2.58  & 0.77 $\pm$ 0.26 & 1646.47 $\pm$ 49.95 & 0.42 $\pm$ 0.84   \\
\textsc{SafetyHumanoidVelocity-v1} & 6579.26 $\pm$ 55.70 & 3.76 $\pm$ 3.61 & 6407.95 $\pm$ 254.06 &7.38 $\pm$ 11.34 & 6321.45 $\pm$ 35.73  & 0.00 $\pm$ 0.00 & 6332.14 $\pm$ 89.86 & 0.00 $\pm$ 0.00   \\
\textsc{SafetySwimmerVelocity-v1} & 91.05 $\pm$ 62.68 & 19.12 $\pm$ 8.33 & 69.75 $\pm$ 46.52 & 20.48 $\pm$ 9.13 & 33.02 $\pm$ 7.26  & 24.23 $\pm$ 0.54 & 39.24 $\pm$ 5.01 & 23.20 $\pm$ 0.48   \\
\textsc{SafetyWalker2dVelocity-v1} & 2183.43 $\pm$ 1300.69 & 14.12 $\pm$ 10.28 & 2707.75 $\pm$ 980.56 & 9.60 $\pm$ 8.94 & 2195.57 $\pm$ 1046.29  & 7.63 $\pm$ 10.44 & 2079.64 $\pm$ 1028.73 & 13.74 $\pm$ 15.94   \\
\textsc{SafetyCarGoal1-v0} & 10.60 $\pm$ 2.51 & 30.66 $\pm$ 7.53 & 25.49 $\pm$ 1.31 & 28.92 $\pm$ 7.66 & 17.92 $\pm$ 1.54  & 21.60 $\pm$ 0.83 & 22.09 $\pm$ 3.07 & 17.97 $\pm$ 1.35   \\
\textsc{SafetyCarButton1-v0} & -1.36 $\pm$ 0.68 & 14.62 $\pm$ 9.40 & -0.31 $\pm$ 0.49 & 15.24 $\pm$ 17.01 & 4.47 $\pm$ 1.12  & 25.00 $\pm$ 0.00 & 4.34 $\pm$ 0.72 & 25.00 $\pm$ 0.00   \\
\textsc{SafetyCarGoal2-v0} & 0.13 $\pm$ 1.11 & 23.50 $\pm$ 1.22  & 1.77 $\pm$ 1.20 & 17.43 $\pm$ 12.13 & 6.59 $\pm$ 0.58  & 25.00 $\pm$ 0.00 & 7.12 $\pm$ 4.06 & 23.37 $\pm$ 1.35   \\
\textsc{SafetyCarButton2-v0} & -1.59 $\pm$ 0.70 & 39.97 $\pm$ 26.91 & -2.95 $\pm$ 4.03 & 27.90 $\pm$ 6.37 & 4.86 $\pm$ 1.57  & 25.00 $\pm$ 0.00 & 5.07 $\pm$ 1.24 & 25.00 $\pm$ 0.00   \\
\textsc{SafetyPointGoal1-v0} & 8.43 $\pm$ 3.43 & 25.74 $\pm$ 7.83 & 19.24 $\pm$ 3.94 & 21.38 $\pm$ 6.96 & 16.03 $\pm$ 8.60  & 19.17 $\pm$ 9.42 & 16.31 $\pm$ 6.99 & 22.10 $\pm$ 6.13  \\
\textsc{SafetyPointButton1-v0} & 1.18 $\pm$ 1.02 & 29.42 $\pm$ 12.10 & 6.40 $\pm$ 1.43 & 27.90 $\pm$ 13.27 & 7.48 $\pm$ 8.47  & 24.27 $\pm$ 3.95 & 9.52 $\pm$ 7.86 & 25.00 $\pm$ 0.00   \\
\textsc{SafetyPointGoal2-v0} & -0.56 $\pm$ 0.06 & 48.43 $\pm$ 40.55 & 1.67 $\pm$ 1.43 & 23.50 $\pm$ 11.17 & 6.09 $\pm$ 5.03  & 25.00 $\pm$ 0.00 & 8.62 $\pm$ 7.13 & 25.00 $\pm$ 0.00   \\
\textsc{SafetyPointButton2-v0} & 0.42 $\pm$ 0.63 & 28.87 $\pm$ 11.27 & 1.00 $\pm$ 1.00 & 30.00 $\pm$ 9.50 & 6.94 $\pm$ 4.47  & 25.00 $\pm$ 0.00  & 8.35 $\pm$ 10.44 & 25.00 $\pm$ 0.00  \\
\bottomrule
\end{tabular}
}

\caption{On-Policy Safe Algorithms.\\ \hspace*{\fill} \\}
\label{on-safe}
\end{subtable}

\end{table}

\pagebreak

\begin{table}[htb]
\ContinuedFloat

\begin{subtable}{\linewidth}\centering
\resizebox{\columnwidth}{!}{
\begin{tabular}{@{}l|cc|cc|cc@{}}\toprule
    & \multicolumn{2}{c|}{\textbf{DDPG}} & \multicolumn{2}{c|}{\textbf{TD3}} & \multicolumn{2}{c}{\textbf{SAC}} \\
\midrule
\textbf{Environment} \hfill  & \textbf{Reward} & \textbf{Cost} & \textbf{Reward}  & \textbf{Cost}  & \textbf{Reward} & \textbf{Cost} \\ \midrule
\textsc{SafetyAntVelocity-v1} & 860.86 $\pm$ 198.03 & 234.80 $\pm$ 40.63 & 5246.86 $\pm$ 580.50 & 912.90 $\pm$ 93.73 & 5456.31 $\pm$ 156.04  & 943.10 $\pm$ 47.51   \\
\textsc{SafetyHalfCheetahVelocity-v1} & 11377.10 $\pm$ 75.29 & 980.93 $\pm$ 1.05 & 11246.12 $\pm$ 488.62 & 981.27 $\pm$ 0.31 & 11488.86 $\pm$ 513.09  & 981.93 $\pm$ 0.33   \\
\textsc{SafetyHopperVelocity-v1} & 1462.56 $\pm$ 591.14 & 429.17 $\pm$ 220.05 & 3404.41 $\pm$ 82.57 & 973.80 $\pm$ 4.92 & 3537.70 $\pm$ 32.23  & 975.23 $\pm$ 2.39   \\
\textsc{SafetyHumanoidVelocity-v1} & 1537.39 $\pm$ 335.62 & 48.79 $\pm$ 13.06 & 5798.01 $\pm$ 160.72 & 255.43 $\pm$ 437.13 & 6039.77 $\pm$ 167.82  & 41.42 $\pm$ 49.78   \\
\textsc{SafetySwimmerVelocity-v1} & 139.39 $\pm$ 11.74 & 200.53 $\pm$ 43.28 & 98.39 $\pm$ 32.28 & 115.27 $\pm$ 44.90 & 46.44 $\pm$ 1.23  & 40.97 $\pm$ 0.47   \\
\textsc{SafetyWalker2dVelocity-v1} & 1911.70 $\pm$ 395.97 & 318.10 $\pm$ 71.03 & 3034.83 $\pm$ 1374.72 & 606.47 $\pm$ 337.33 & 4419.29 $\pm$ 232.06  & 877.70 $\pm$ 8.95   \\
\bottomrule
\end{tabular}
}

\caption{Off-Policy Base Algorithms.\\ \hspace*{\fill} \\}
\label{off-base}
\end{subtable}

\begin{subtable}{\linewidth}\centering
\resizebox{\columnwidth}{!}{
\begin{tabular}{@{}l|cc|cc|cc@{}}\toprule
    & \multicolumn{2}{c|}{\textbf{DDPGLag}} & \multicolumn{2}{c|}{\textbf{TD3Lag}} & \multicolumn{2}{c}{\textbf{SACLag}} \\
\midrule
\textbf{Environment} \hfill  & \textbf{Reward} & \textbf{Cost} & \textbf{Reward}  & \textbf{Cost}  & \textbf{Reward} & \textbf{Cost} \\ \midrule
\textsc{SafetyAntVelocity-v1} & 1271.48 $\pm$ 581.71 & 33.27 $\pm$ 13.34 & 1944.38 $\pm$ 759.20 & 63.27 $\pm$ 46.89 & 1897.32 $\pm$ 1213.74  & 5.73 $\pm$ 7.83   \\
\textsc{SafetyHalfCheetahVelocity-v1} & 2743.06 $\pm$ 21.77 & 0.33 $\pm$ 0.12 & 2741.08 $\pm$ 49.13 & 10.47 $\pm$ 14.45 & 2833.72 $\pm$ 3.62  & 0.00 $\pm$ 0.00   \\
\textsc{SafetyHopperVelocity-v1} & 1093.25 $\pm$ 81.55 & 15.00 $\pm$ 21.21 & 928.79 $\pm$ 389.48 & 40.67 $\pm$ 30.99 & 963.49 $\pm$ 291.64  & 20.23 $\pm$ 28.47   \\
\textsc{SafetyHumanoidVelocity-v1} & 2059.96 $\pm$ 485.68 & 19.71 $\pm$ 4.05 & 5751.99 $\pm$ 157.28 & 10.71 $\pm$ 23.60 & 5940.04 $\pm$ 121.93  & 17.59 $\pm$ 6.24   \\
\textsc{SafetySwimmerVelocity-v1} & 13.18 $\pm$ 20.31 & 28.27 $\pm$ 32.27 & 15.58 $\pm$ 16.97 & 13.27 $\pm$ 17.64 & 11.03 $\pm$ 11.17  & 22.70 $\pm$ 32.10   \\
\textsc{SafetyWalker2dVelocity-v1} & 2238.92 $\pm$ 400.67 & 33.43 $\pm$ 20.08 & 2996.21 $\pm$ 74.40 & 22.50 $\pm$ 16.97 & 2676.47 $\pm$ 300.43  & 30.67 $\pm$ 32.30   \\
\bottomrule
\end{tabular}
}

\caption{Off-Policy Safe Algorithms.\\ \hspace*{\fill} \\}
\label{off-safe}
\end{subtable}

\begin{subtable}{\linewidth}\centering
\resizebox{\columnwidth}{!}{
\begin{tabular}{@{}l|cc|cc|cc@{}}\toprule
    & \multicolumn{2}{c|}{\textbf{PETS}} & \multicolumn{2}{c|}{\textbf{LOOP}} & \multicolumn{2}{c}{\textbf{SafeLOOP}} \\
\midrule
\textbf{Environment} \hfill  & \textbf{Reward} & \textbf{Cost} & \textbf{Reward}  & \textbf{Cost} & \textbf{Reward} & \textbf{Cost}  \\ \midrule
\textsc{SafetyCarGoal1-v0} & 33.07 $\pm$ 1.33 & 61.20 $\pm$ 7.23 & 25.41 $\pm$ 1.23 & 62.64 $\pm$ 8.34  & 22.09 $\pm$ 0.30 & 0.16 $\pm$ 0.15   \\
\textsc{SafetyPointGoal1-v0} & 27.66 $\pm$ 0.07 & 49.16 $\pm$ 2.69 & 25.08 $\pm$ 1.47 & 55.23 $\pm$ 2.64 & 22.94 $\pm$ 0.72 & 0.04 $\pm$ 0.07   \\
\bottomrule
\end{tabular}
}
\end{subtable}

\begin{subtable}{\linewidth}\centering
\resizebox{\columnwidth}{!}{
\begin{tabular}{@{}l|cc|cc|cc@{}}\toprule
    & \multicolumn{2}{c|}{\textbf{CCEPETS}} & \multicolumn{2}{c|}{\textbf{RCEPETS}} & \multicolumn{2}{c}{\textbf{CAPPETS}}\\
\midrule
\textbf{Environment} \hfill  & \textbf{Reward} & \textbf{Cost} & \textbf{Reward}  & \textbf{Cost} & \textbf{Reward} & \textbf{Cost} \\ \midrule
\textsc{SafetyCarGoal1-v0} & 27.60 $\pm$ 1.21 & 1.03 $\pm$ 0.29 & 29.08 $\pm$ 1.63 & 1.02 $\pm$ 0.88 & 23.33 $\pm$ 6.34 & 0.48 $\pm$ 0.17   \\
\textsc{SafetyPointGoal1-v0} & 24.98 $\pm$ 0.05 & 1.87 $\pm$ 1.27 & 25.39 $\pm$ 0.28 & 2.46 $\pm$ 0.58   & 9.45 $\pm$ 8.62 & 0.64 $\pm$ 0.77  \\
\bottomrule
\end{tabular}
}

\caption{Model-based Algorithms.\\ \hspace*{\fill} \\}
\label{mb}
\end{subtable}

\begin{subtable}{\linewidth}\centering
\resizebox{\columnwidth}{!}{
\begin{tabular}{@{}l|cc|cc|cc|cc@{}}\toprule
    & \multicolumn{2}{c|}{\textbf{VAE-BC}} & \multicolumn{2}{c|}{\textbf{C-CRR}} & \multicolumn{2}{c|}{\textbf{COptiDICE}} & \multicolumn{2}{c}{\textbf{BCQ-Lag}}\\
\midrule
\textbf{Environment} \hfill  & \textbf{Reward} & \textbf{Cost} & \textbf{Reward}  & \textbf{Cost}  & \textbf{Reward} & \textbf{Cost} & \textbf{Reward}  & \textbf{Cost}\\ \midrule
\textsc{SafetyCarCircle1-v0} & 17.31 $\pm$ 0.33 & 85.53 $\pm$ 11.33 & 15.74 $\pm$ 0.42 & 48.38 $\pm$ 10.31 & 15.58$\pm$ 0.37 & 49.42 $\pm$ 8.70 & 17.10 $\pm$ 0.84 & 77.54 $\pm$ 14.07 \\
\textsc{SafetyCarRun0-v0} & 821.21 $\pm$ 3.27 & 45.68 $\pm$ 8.25 & 635.91 $\pm$ 69.72 & 103.91 $\pm$ 50.68 & 807.24 $\pm$ 7.12 & 53.92 $\pm$ 11.41 & 813.58 $\pm$ 5.52 & 55.86 $\pm$ 8.43 \\
\textsc{SafetyPointCircle1-v0} & 40.23 $\pm$ 0.75 & 41.25 $\pm$ 10.12 & 40.66 $\pm$ 0.88 & 49.90 $\pm$ 10.81 & 40.98 $\pm$ 0.89 & 70.40 $\pm$ 12.14 & 42.94 $\pm$ 1.04 & 85.37 $\pm$ 23.41 \\
\textsc{SafetyPointRun0-v0} & 813.47 $\pm$ 4.40 & 137.38 $\pm$ 26.02 & 801.77 $\pm$ 1.69 & 63.05 $\pm$ 5.04 & 776.94 $\pm$ 3.80 & 51.22 $\pm$ 12.15 & 828.95 $\pm$ 0.46 & 46.01 $\pm$ 3.02 \\
\bottomrule
\end{tabular}
}

\caption{Offline Algorithms.\\ \hspace*{\fill} \\}
\label{offline}
\end{subtable}
\caption{
The performance of \omni algorithms, encompassing both reward and cost, was assessed within the \sg environments. It is crucial to highlight that all on-policy algorithms in \autoref{on-base} and \autoref{on-safe} underwent evaluation following 1e7 training steps. In addition, the performance of on-policy \autoref{on-safe}, off-policy \autoref{off-safe}, and offline \autoref{offline} algorithms was obtained under the experimental setting of \texttt{cost\_limit=25.00}, except for the model-based \autoref{mb} algorithms with \texttt{cost\_limit=1.00}. During experimentation, it was observed that off-policy algorithms did not violate safety constraints in \texttt{SafetyHumanoidVeloicty-v1}. This observation suggests that the agent may not have fully learned to run within 1e6 steps; consequently, the 3e6 results were utilized in off-policy \texttt{SafetyHumanoidVeloicty-v1}. With this exception in consideration, all off-policy, offline and model-based algorithms were evaluated after 1e6 training steps, as demonstrated in \autoref{off-base}, \autoref{off-safe}, \autoref{mb}, \autoref{offline}.
}
\label{performance}
\end{table}



\clearpage

\section{Experiment Grid of \omni} \label{pseudocode}

\begin{figure}[htb]
  \centering
  \includegraphics[width=1.0\linewidth]{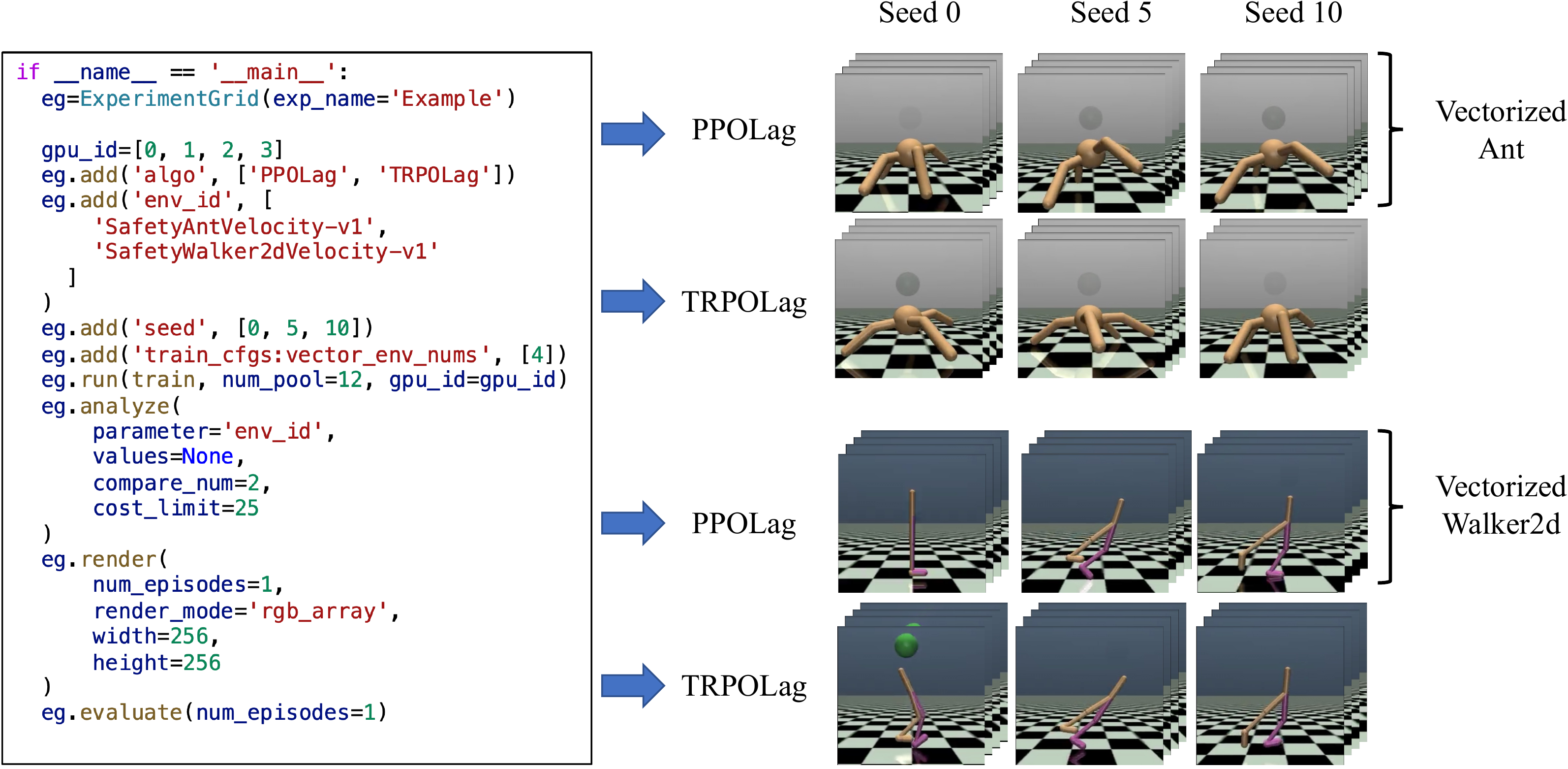}
  \caption{\omni's \texttt{Experiment Grid}. The left side of the figure displays the main function of the \texttt{run\_experiment\_grid.py} file,  while the right side shows the status of the \texttt{Experiment Grid} execution. In this example, three distinct random seeds are selected for the \texttt{SafetyAntVelocity-v1} and \texttt{SafetyWalker2dVelocity-v1}, then the PPO-Lag and TRPO-Lag algorithms are executed.}
  \label{expgrid}
\end{figure}

In the context of reinforcement learning experiments, it is imperative to assess the performance of various algorithms across multiple environments. However, the inherent influence of randomness necessitates repeated evaluations employing distinct random seeds. To tackle this challenge, \omni introduces an \texttt{Experiment Grid}, facilitating simultaneous initiation of multiple experimental sets. Researchers are merely required to pre-configure the experimental parameters, subsequently executing multiple experiments sets in parallel via a single file. An exemplification of this process can be found in \autoref{expgrid}.

\begin{figure}[htb]
  \centering
  \begin{subfigure}{0.49\linewidth}
    \centering
    \includegraphics[width=\linewidth]{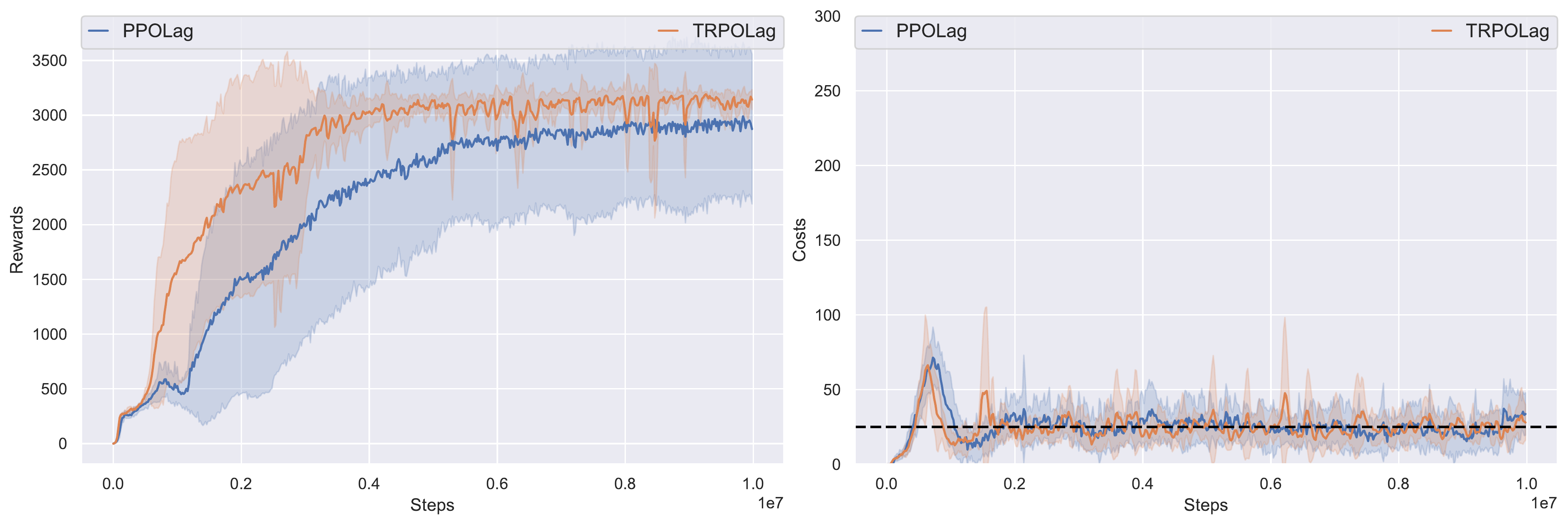}
    \caption{Walker2d}
  \end{subfigure}
  \begin{subfigure}{0.49\linewidth}
    \centering
    \includegraphics[width=\linewidth]{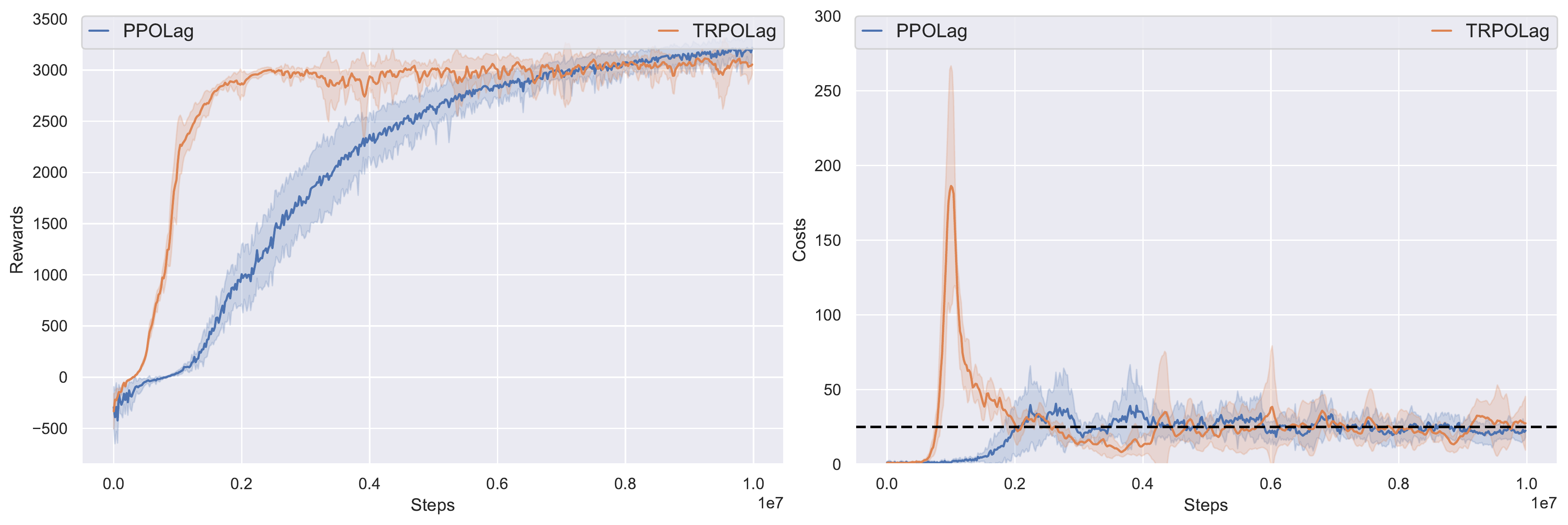}
    \caption{Ant}
  \end{subfigure}
\caption{Analysis of the \autoref{expgrid} example experiment results. The blue lines are the results from PPO-Lag while the orange ones are TRPO-Lag.
The solid line in the figure represents the mean of multiple random seeds, while the shadow represents the standard deviation among 0, 5, 10 random seeds. }
  \label{gridresult}
\end{figure}

The \texttt{run\_experiment\_grid.py} script executes experiments in parallel based on user-configured parameters and generates corresponding graphs of the experimental results. In the example presented in \autoref{expgrid}, we specified that the script should draw curves based on different environments and obtained the training curves of PPO-Lag and TRPO-Lag in \texttt{SafetyAntVelocity-v1} and \texttt{SafetyWalker2dVelocity-v1}, where seeds have been grouped.

\begin{figure}[htb]
  \centering
  \includegraphics[width=1.0\linewidth]{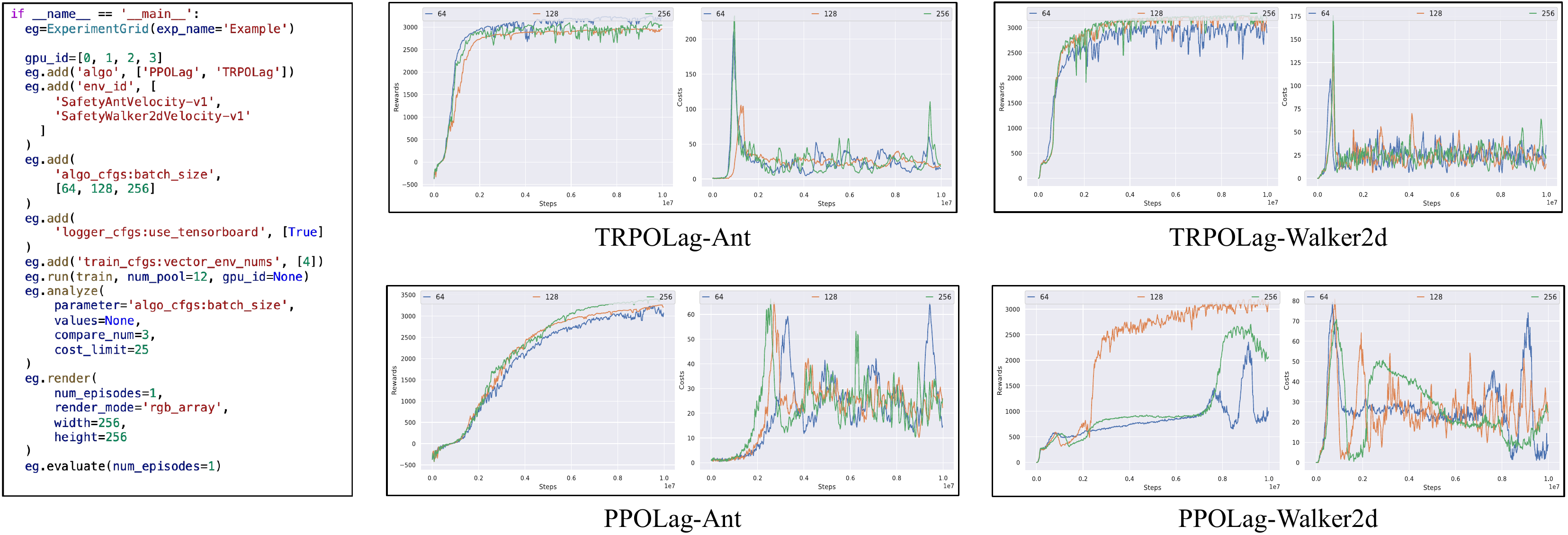}
  \caption{An example of how the \texttt{Experiment Grid} can be utilized for parameters tuning. In this particular example, we set the \texttt{batch\_size} in the \texttt{algo\_cfgs} to 64, 128, and 256. Then we ran multiple experiments using the \texttt{Experiment Grid}, finally used \texttt{statistics tools} to analyze the impact of the \texttt{batch\_size} on the performance of the algorithm. Note that different colors denote different \texttt{batch\_size}. The results showed that the \texttt{batch\_size} had a significant effect on the performance of the algorithm, and the optimal \texttt{batch\_size} was found to be 128. The \texttt{Experiment Grid} enabled us to efficiently explore the effect of different parameter values on the algorithm's performance.}
  \label{analyze}
\end{figure}

Moreover, combined with \omni \texttt{statistics
tools}, the \texttt{Experiment Grid} is a powerful tool for parameter tuning. As illustrated in \autoref{analyze}, we utilized the \texttt{Experiment Grid} to explore the impact of \texttt{batch\_size} on the performance of PPO-Lag and TRPO-Lag in \texttt{SafetyWalker2dVelocity-v1} and \texttt{SafetyAntVelocity-v1}, then used \texttt{statistics tools} to analyze the experiment results. It is obvious that the \texttt{batch\_size} has a significant influence on the performance of PPO-Lag in \texttt{SafetyAntVelocity-v1}, and the optimal \texttt{batch\_size} is 128. Obtaining this conclusion requires repeating the experiment multiple times, and the \texttt{Experiment Grid} significantly expedites the process.

\clearpage

\section{The Documentation of \omni}

\begin{figure}[htb]
  \centering
  \includegraphics[width=1.0\linewidth]{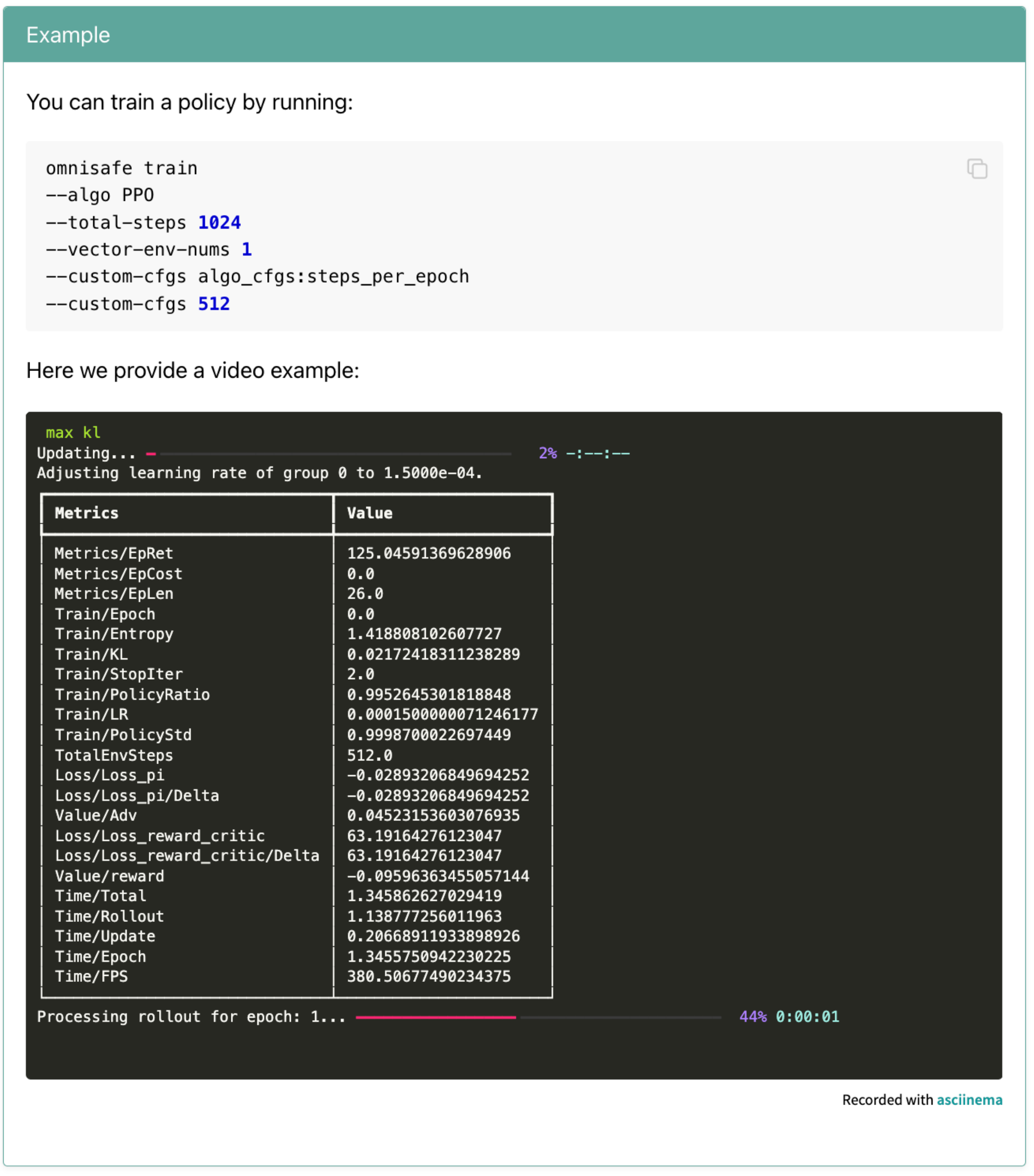}
  \caption{A video example of a single algorithm training. The upper part shows the command to start training using PPO with \omni in the default environment \texttt{SafetyPointGoal1-v0}, while the lower part shows the information printed on the terminal after executing this command.}
  \label{video}
\end{figure}

The \omni documentation provides quick start video examples to help new users easily get started with the platform. These videos cover a range of topics, including algorithm training and evaluation, running the \texttt{Experiment Grid}, and accessing help. \autoref{video} demonstrates the execution commands and video screenshots for single algorithm training, providing a clear and straightforward guide for users to follow along and begin using \texttt{Experiment Grid} quickly.

\begin{figure}[htb]
  \centering
  \includegraphics[width=1.0\linewidth]{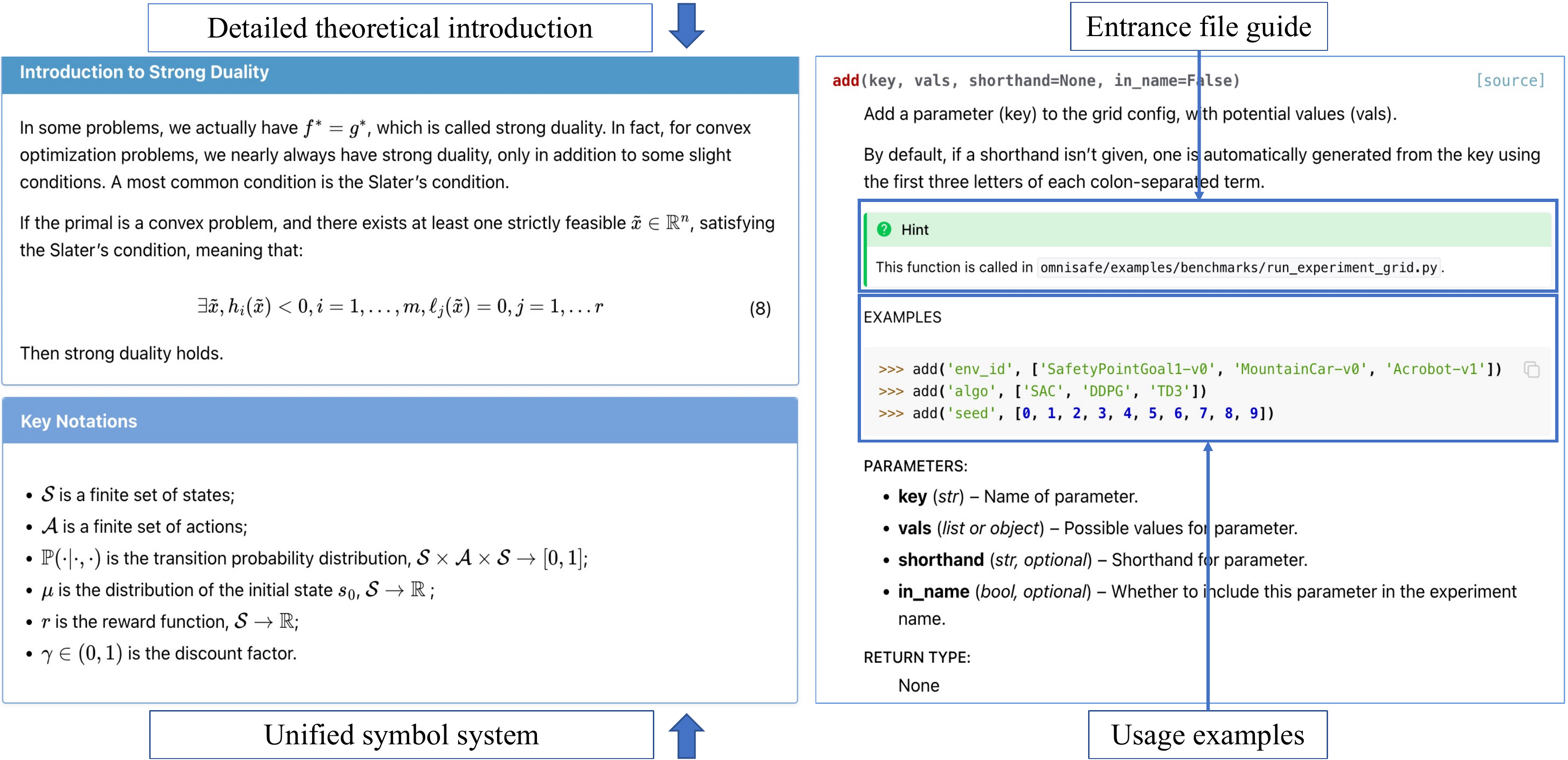}
  \caption{The left half of the figure excerpted a detailed explanation of the Lagrange duality theorem and part of the symbol system explanation from the \omni documentation. The right half, taking the function \texttt{add} as an example, demonstrates how the API documentation of \omni enables users to quickly understand the usage of the framework. The convenience of \omni documentation includes: (a). A unified symbol system; (b). Easy-to-understand introductory guidance; (c). Detailed API documentation and usage examples.}
  \label{docs_guide}
\end{figure}

The documentation of \omni has the characteristic of being user-friendly, making it easy for users to get started. Since algorithms in the SafeRL field are often based on rigorous mathematical derivations, understanding these algorithms requires prerequisite mathematical knowledge, such as linear algebra and optimization theory. This can make it difficult for many researchers to get started. The \omni documentation provides introductory-level mathematical theory tutorials, allowing researchers to quickly understand the prerequisite mathematical knowledge of SafeRL algorithms. At the same time, many existing SafeRL algorithms often have their own symbol systems, using different symbols to represent the same meaning, which can easily confuse beginners. The \omni documentation unifies the symbol system of the SafeRL field, enabling readers to learn classic SafeRL algorithms coherently, from CPO\citep{cpo2017}, PCPO\citep{pcpo2020} to FOCOPS\citep{focops2020}, and so on. In addition, to facilitate researchers' quick understanding of the functions defined in \omni, we provide usage examples and detailed introduction in the API documentation, making every effort to serve the convenience of users.

\begin{figure}[htb]
  \centering
  \includegraphics[width=1.0\linewidth]{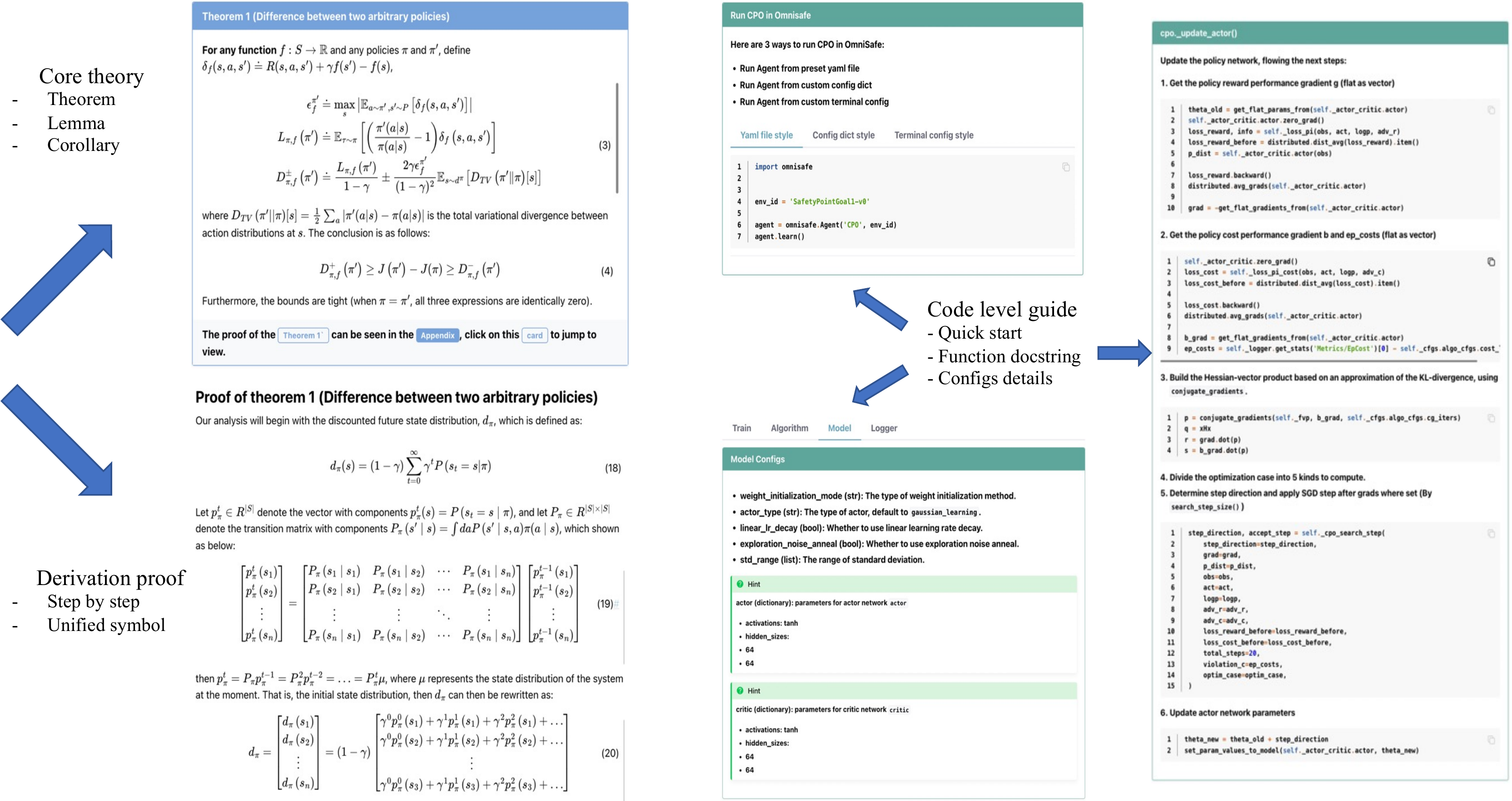}
  \caption{Excerpt of the core part of the \omni's CPO documentation. The left half illustrates the core theory and related proofs of CPO, while the right half introduces the specific implementation of the CPO algorithm from three levels: entry files, parameters explanations, and core functions. All algorithm tutorials provided by \omni are arranged in a hierarchical structure of theory, proof, and connection between theory and code, which is convenient for beginners to get started quickly. The main features of \omni documentation includes: (a). Detailed and rigorous theoretical derivation; (b). Simple and clear code tutorial; (c). Linking theory to code narrative style.}
  \label{docs_cpo}
\end{figure}

\autoref{docs_cpo} highlights some of the key features of \omni's documentation, using CPO \citep{cpo2017} as an example. The left-hand side of the \autoref{docs_cpo} displays the theoretical part of CPO, which is presented in the form of cards with clickable hyperlinks. This allows readers to easily access the theoretical proof section of CPO by clicking on the corresponding card. On the right-hand side of the page, \omni provides a clear and simple code tutorial on how to implement CPO, starting from the code level. The tutorial includes the command to run the CPO algorithm in \omni, an explanation of the parameters related to CPO, and a step-by-step interpretation of the core functions of CPO. This linking of theory to code narrative style enhances the understanding of the CPO algorithm for users.

\clearpage

\section{Features of \omni}
\omni transcends its role as a mere infrastructure for expediting SafeRL research, functioning concurrently as a standardized and user-friendly reinforcement learning framework. We compared the features of \omni with popular open-source reinforcement learning benchmarks such as \ts, \stable, \clean\footnote{The Project URL: \url{https://github.com/vwxyzjn/cleanrl}}, \safety\footnote{The Project URL: \url{https://github.com/openai/safety-starter-agents}}, \ray\footnote{The Project URL: \url{https://www.ray.io/}} etc. to ensure that \omni adheres to the same standards as these well-established libraries\footnote{All results in \autoref{compare_with_repo} are accurate as of May 16, 2023. If you find any discrepancies between these data, please consider the latest results as accurate.}.

\begin{table*}[htb] \centering
\begin{small}
\resizebox{\columnwidth}{!}{%
\begin{tabular}{@{}l|c|c|c|c|c|c@{}}\toprule
  \textbf{Features} \hfill & \textbf{\omni}  & \textbf{\ts}  & \textbf{\stable} & \textbf{\clean} & \textbf{\ray}  & \textbf{\safety}  \\
\midrule 
\textbf{Ipython / Notebook}  & \cmark  & \cmark & \cmark & \cmark  & \cmark & \xmark\\
\textbf{API Documentation}  & \cmark  & \cmark & \cmark & \cmark  & \cmark & \xmark\\
\textbf{Algorithm Tutorial}  & \cmark  & \cmark & \cmark & \cmark  & \cmark & \xmark\\
\textbf{Test Coverage}  & $97\%$  & $91\%$ & $96\%$ & - & - & -\\
\textbf{Type Hints}  & \cmark  & \cmark & \cmark & \cmark & \cmark & \cmark\\
\textbf{TensorBoard Support}  & \cmark  & \cmark & \cmark & \cmark  & \cmark & \xmark\\
\textbf{WandB Report} & \cmark  & \cmark & \cmark & \cmark  & \cmark & \xmark\\
\textbf{PEP8 Code Style} & \cmark  & \cmark & \cmark & \cmark  & \cmark & \cmark\\
\textbf{Statistics Tools} & \cmark  & \cmark & \cmark & \cmark & \cmark & \xmark\\
\textbf{Video Examples} & \cmark  & \cmark & \cmark & \cmark  & \cmark & \xmark\\
\textbf{Command Line Interface} & \cmark  & \cmark & \cmark & \cmark & \cmark & \xmark\\
\midrule
\bottomrule
\end{tabular}
}
\end{small}
\caption{Comparsion of \omni to a representative subset of RL or SafeRL libraries.}
\label{compare_with_repo}
\end{table*}

The complete codebase adheres to the PEP8 style, with each commit undergoing stringent evaluations, such as \texttt{isort}, \texttt{pylint}, \texttt{black}, and \texttt{ruff}. Prior to merging into the main branch, code modifications necessitate approval from a minimum of two reviewers.

\begin{figure}[htb]
  \centering
  \includegraphics[width=1.0\linewidth]{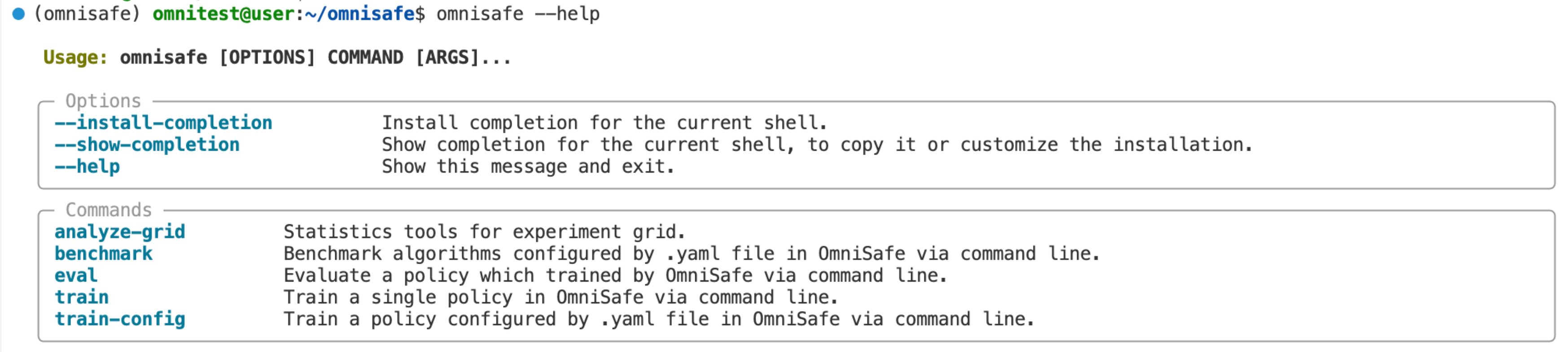}
  \caption{An illustration of the \omni command line interface is provided. Users can view the commands supported by \omni and a brief usage guide by simply typing \texttt{omnisafe --help} in the command line. If a user wants to further understand how to use a specific command, they can obtain additional prompts by using the command \texttt{omnisafe COMMAND --help}, as shown in the \autoref{cli_details}.}
  \label{cli}
\end{figure}

\begin{figure}[htb]
  \centering
  \begin{subfigure}{\linewidth}
    \centering
    \includegraphics[width=0.88\linewidth]{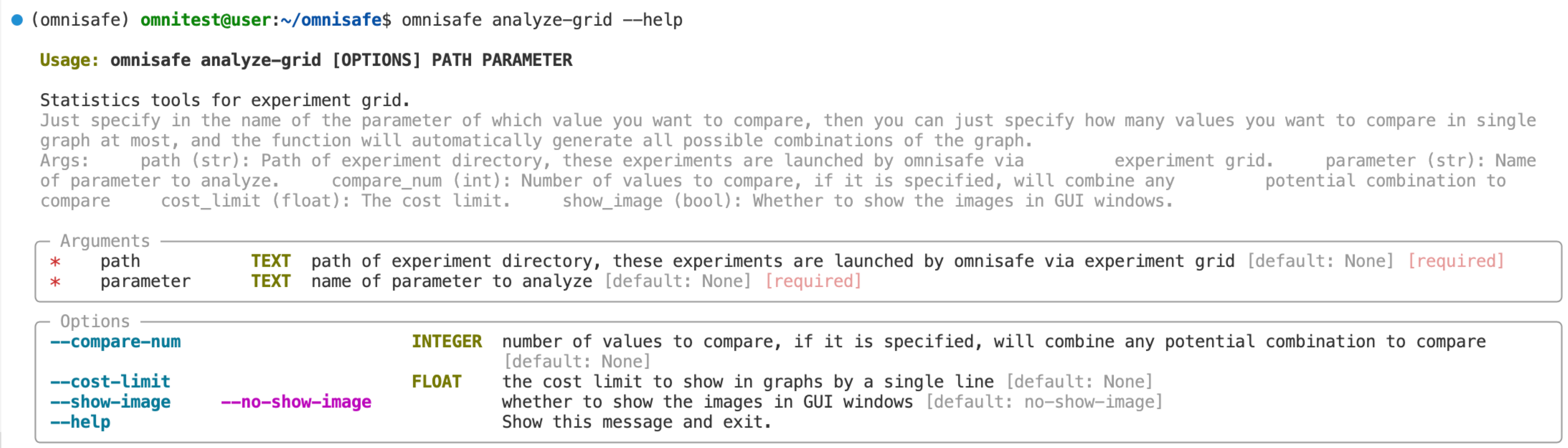}
    \caption{Example of \texttt{omnisafe analyze-grid --help} in command line.}
  \end{subfigure}
  \begin{subfigure}{\linewidth}
    \centering
    \includegraphics[width=0.88\linewidth]{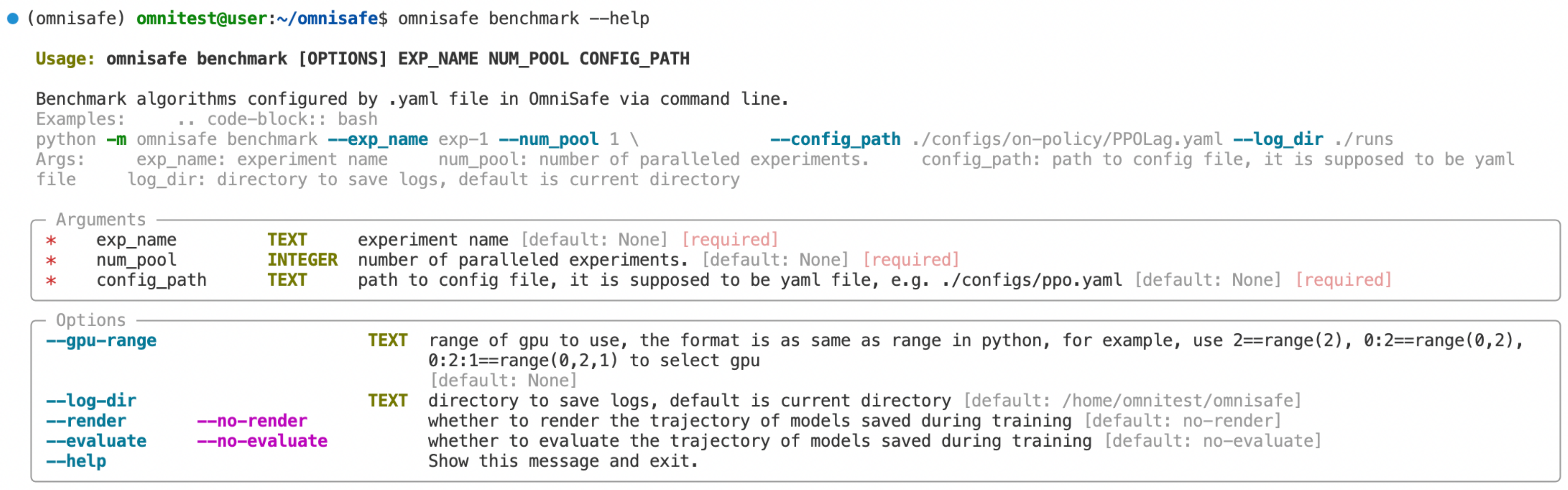}
    \caption{Example of \texttt{omnisafe benchmark --help} in command line.}
  \end{subfigure}
  \begin{subfigure}{\linewidth}
    \centering
    \includegraphics[width=0.88\linewidth]{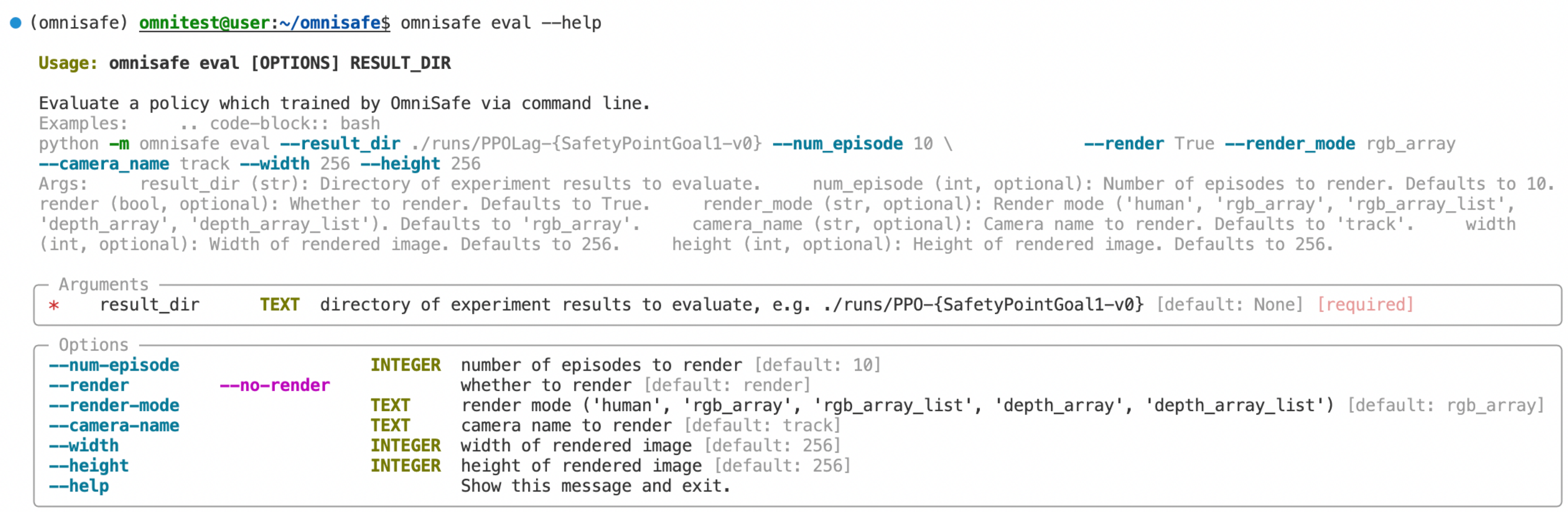}
    \caption{Example of \texttt{omnisafe eval --help} in command line.}
  \end{subfigure}
  \begin{subfigure}{\linewidth}
    \centering
    \includegraphics[width=0.88\linewidth]{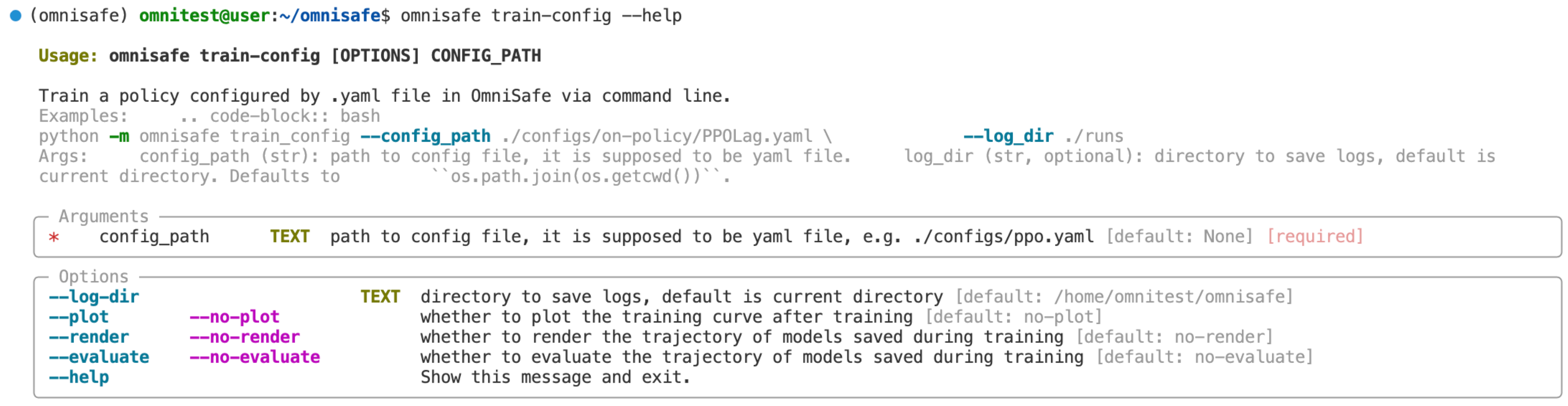}
    \caption{Example of \texttt{omnisafe train-config --help} in command line.}
  \end{subfigure}
\caption{Here are some more details on using \texttt{omnisafe --help} command. Users can input \texttt{omnisafe COMMAND --help} to get help, where \texttt{COMMAND} includes all the items listed in \texttt{Commands} of \autoref{cli}. This feature enables users to swiftly acquire proficiency in executing common operations provided by \omni via command-line, as well as to customize them further to meet their specific requirements.}
  \label{cli_details}
\end{figure}

\omni furnishes users with exhaustive documentation, encompassing algorithm
overviews and API documentation. Seasoned researchers can capitalize on \omni's informative command-line interface, as demonstrated in \autoref{cli} and \autoref{cli_details}, facilitating rapid comprehension of the platform's utilization to expedite their scientific investigations.

\clearpage

\begin{figure}[htb]
  \centering
  \includegraphics[width=0.75\linewidth]{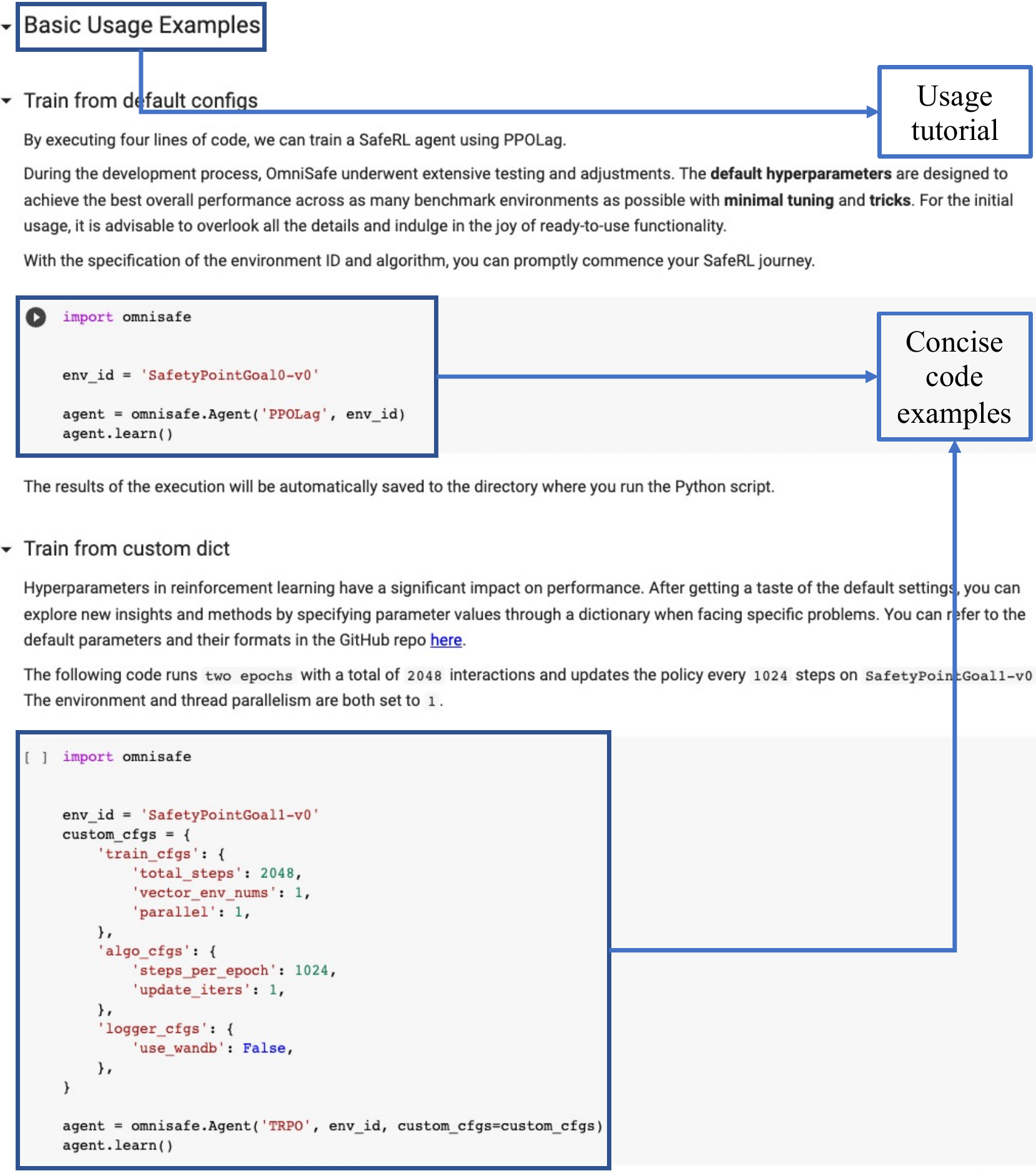}
  \caption{A example demonstrating the colab tutorial provided by \omni for using the \texttt{Experiment Grid}. The tutorial includes detailed descriptions of usage and allows users to try running it then see the results.}
  \label{tutorial}
\end{figure}

\omni includes a tutorial on \texttt{Colab} that provides a step-by-step guide to the training process, as illustrated in \autoref{tutorial}. For those who are new to SafeRL, the tutorial provides an introduction to the underlying theory, and allows for interactive learning of the training procedure. By clicking on \texttt{Colab Tutorial}, users can access it and follow along with the instructions to gain a better understanding of how to use \omni.

\begin{figure}[htb]
  \centering
  \begin{subfigure}{0.48\linewidth}
    \centering
    \includegraphics[width=\linewidth]{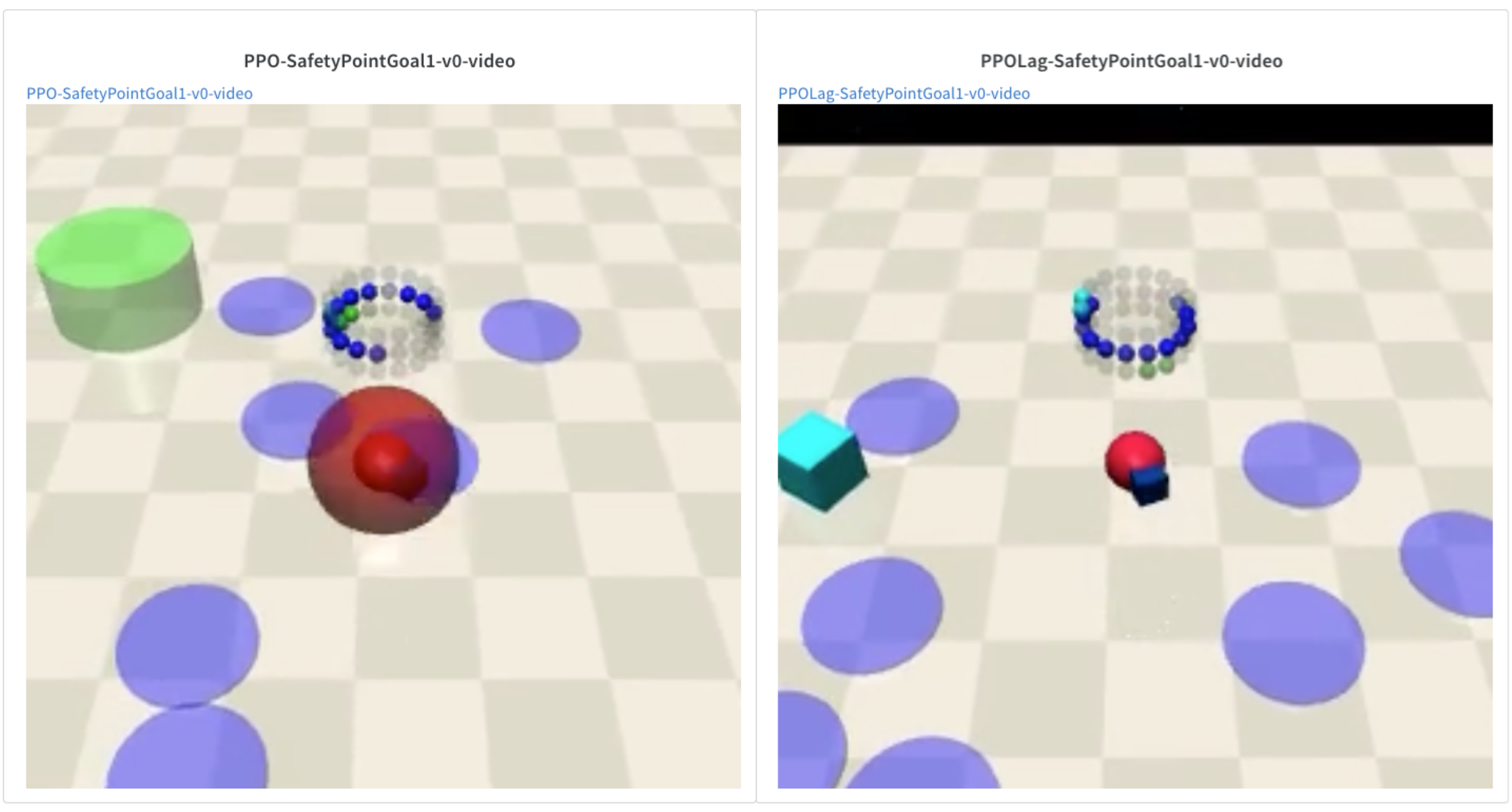}
    \caption{\texttt{SafetyPointGoal1-v0}}
  \end{subfigure}
  \begin{subfigure}{0.48\linewidth}
    \centering
    \includegraphics[width=\linewidth]{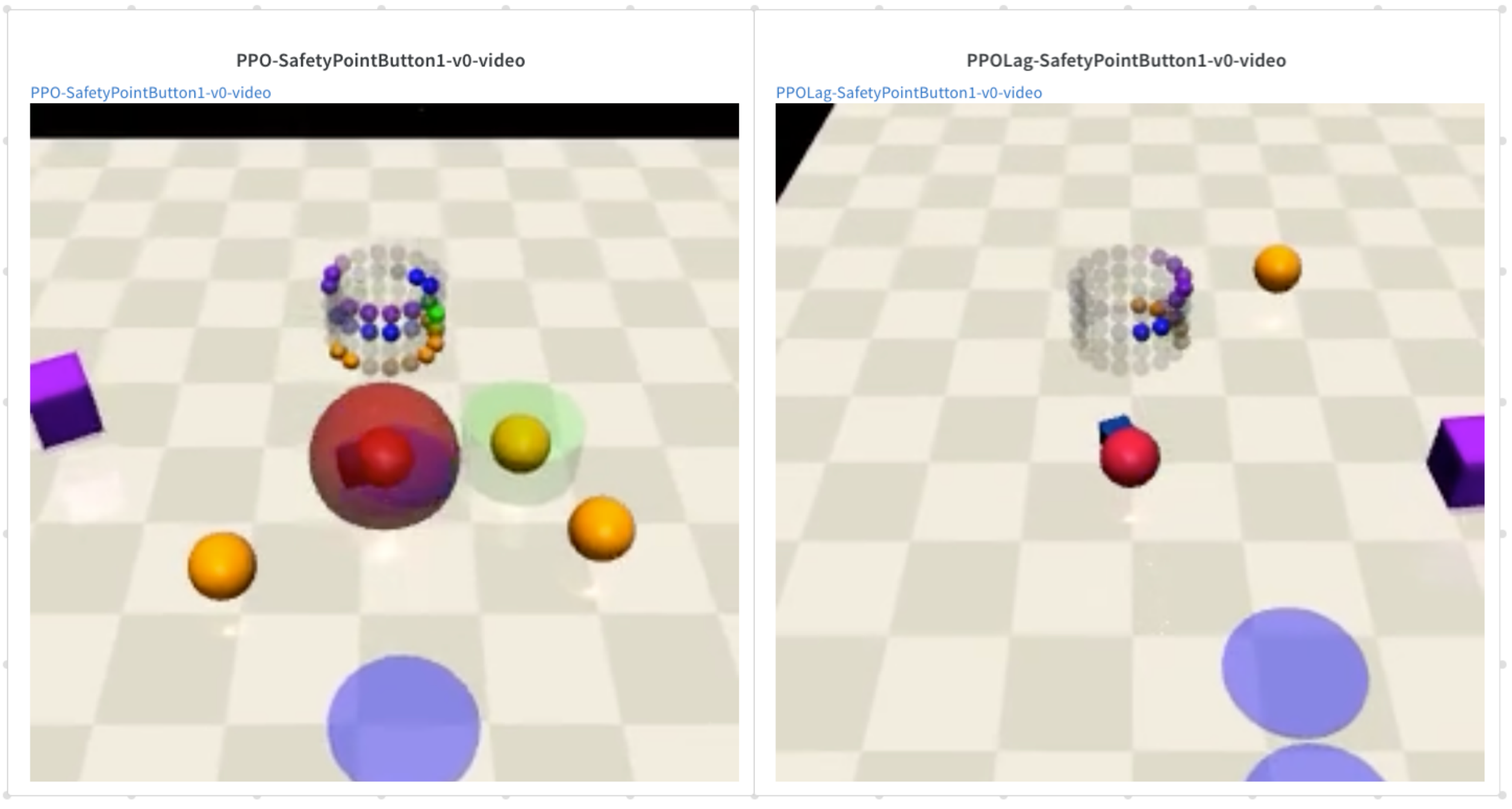}
    \caption{\texttt{SafetyPointButton1-v0}}
  \end{subfigure}
  \begin{subfigure}{0.48\linewidth}
    \centering
    \includegraphics[width=\linewidth]{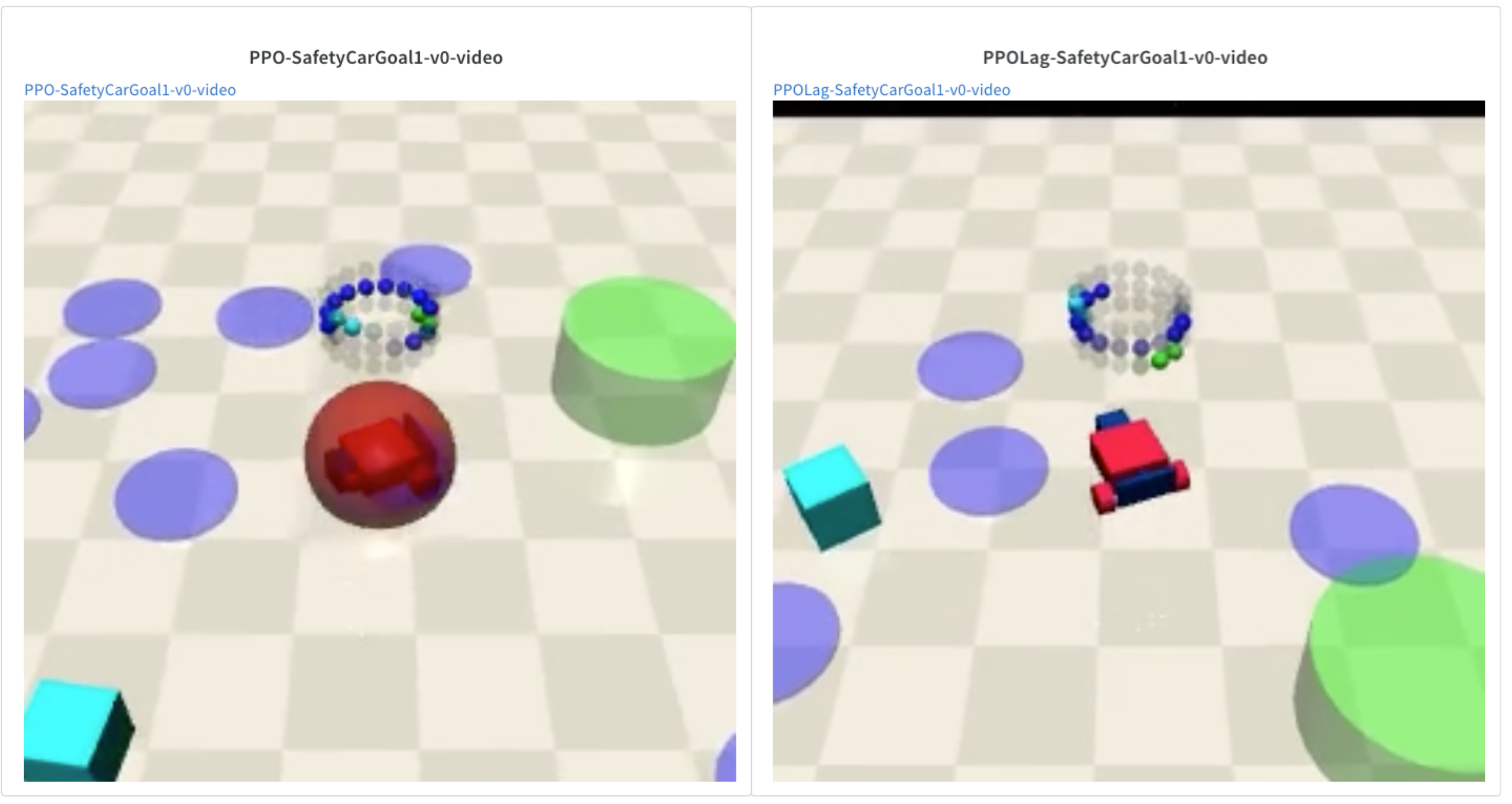}
    \caption{\texttt{SafetyCarGoal1-v0}}
  \end{subfigure}
  \begin{subfigure}{0.48\linewidth}
    \centering
    \includegraphics[width=\linewidth]{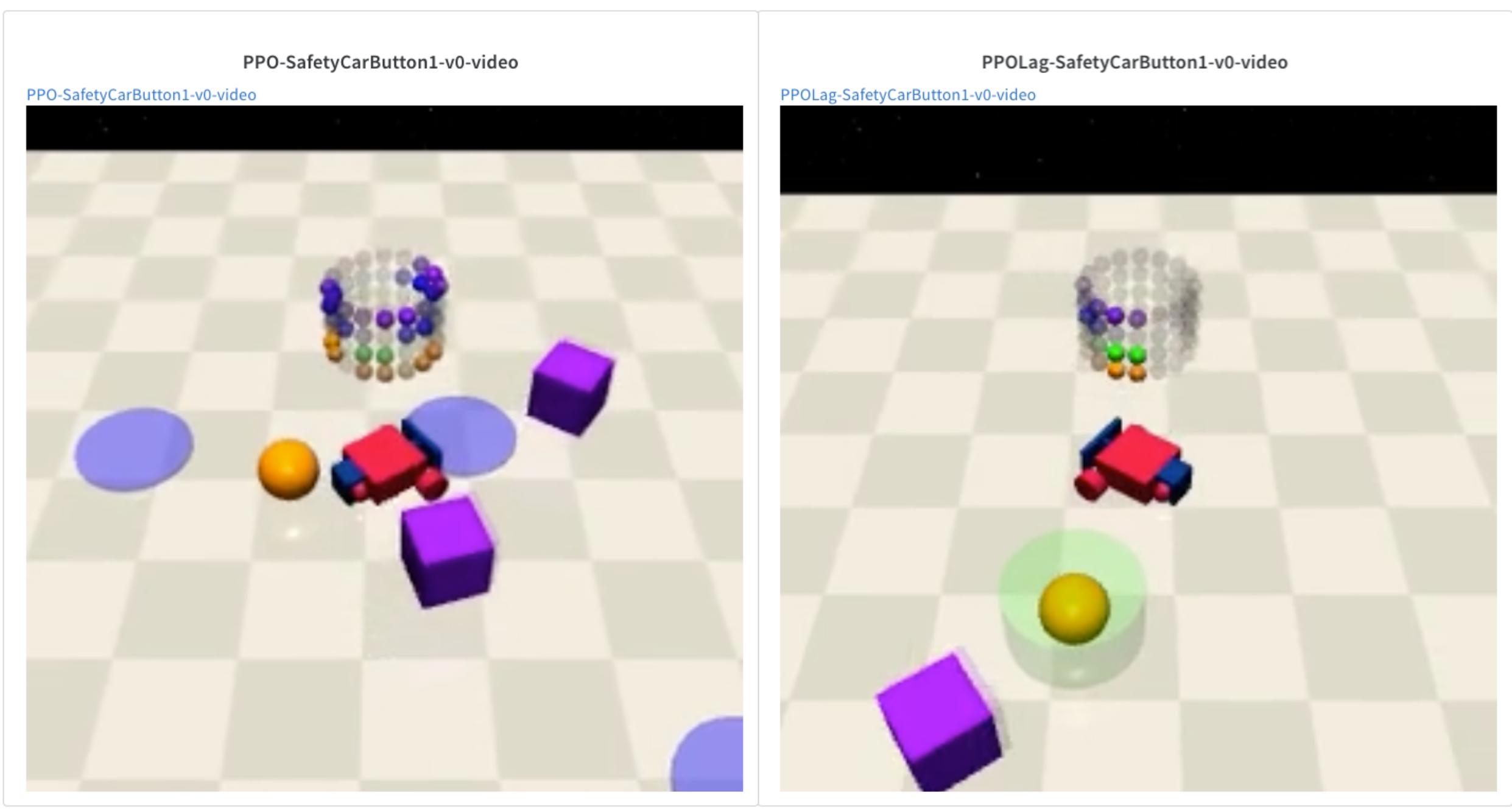}
    \caption{\texttt{SafetyCarButton1-v0}}
  \end{subfigure}
\caption{An exemplification of \omni's \texttt{WandB} reports videos. This example supplies videos of PPO and PPO-Lag in \texttt{SafetyPointGoal1-v0}, \texttt{SafetyPointButton1-v0}, \texttt{SafetyCarGoal1-v0} and \texttt{SafetyCarButton1-v0} environments. The left of each sub-figure is PPO while the right is PPO-Lag. Through these videos, we can intuitively witness the difference between safe and unsafe behavior. This is exactly what \omni pursues: not just the safety of the training curve, but the true safety in a real sense.}
  \label{wandb_video}
\end{figure}

Regarding the experiment execution process, \omni presents an array of tools for analyzing experimental outcomes, encompassing \texttt{WandB}, \texttt{TensorBoard}, and \texttt{statistics tools}. Furthermore, \omni has submitted its experimental benchmark to the \texttt{WandB} report, as depicted in \autoref{wandb_video}. This benchmark furnishes training curves and evaluation demonstrations, serving as a valuable reference for researchers.

\begin{figure}[htb]
    \centering
    \includegraphics[width=1.0\linewidth]{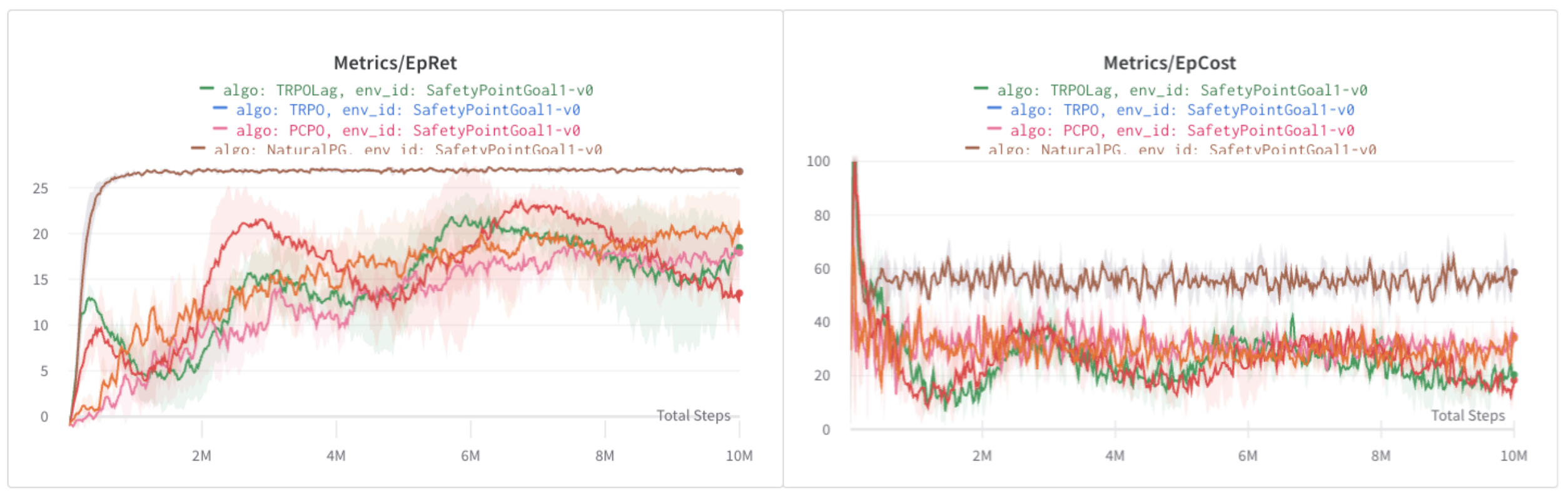}
\caption{An exemplification of \omni's \texttt{WandB} reports training curve in \texttt{SafetyPointGoal1-v0}: The left panel represents the episode reward, and the right panel denotes the episode cost, with both encompassing the performance over 1e7 steps.}
  \label{wandb_curve}
\end{figure}

In addition to videos, we also provide the algorithms performance curves of \omni in \texttt{WandB} reports. An example is provided in \autoref{wandb_curve}. Compared to our publicly available experimental results, \texttt{WandB} reports offer users a more comprehensive view of algorithm performance in some specific environments. With \texttt{WandB} reports, users can easily access insightful information from multiple perspectives, providing them with a more inspiring and informative experience.

\end{document}